\documentclass[sigconf,screen,nonacm]{acmart}
\acmSubmissionID{1282}

\usepackage{booktabs} 

\usepackage[ruled]{algorithm2e} 

\SetAlFnt{\small}
\SetAlCapFnt{\small}
\SetAlCapNameFnt{\small}
\SetAlCapHSkip{0pt}

\copyrightyear{2024}
\acmYear{2024}
\setcopyright{acmlicensed}\acmConference[SIGGRAPH Conference Papers '24]{Special Interest Group on Computer Graphics and Interactive Techniques Conference Conference Papers '24}{July 27-August 1, 2024}{Denver, CO, USA}
\acmBooktitle{Special Interest Group on Computer Graphics and Interactive Techniques Conference Conference Papers '24 (SIGGRAPH Conference Papers '24), July 27-August 1, 2024, Denver, CO, USA}
\acmDOI{10.1145/3641519.3657528}
\acmISBN{979-8-4007-0525-0/24/07}

\usepackage[inline]{enumitem} 
\usepackage{color}
\usepackage{multirow}
\usepackage[makeroom]{cancel}
\usepackage{bm}

\definecolor{turquoise}{cmyk}{0.65,0,0.1,0.3}
\definecolor{purple}{rgb}{0.65,0,0.65}
\definecolor{dark_purple}{rgb}{0.5,0,0.5}
\definecolor{dark_green}{rgb}{0, 0.5, 0}
\definecolor{orange}{rgb}{0.8, 0.6, 0.2}
\definecolor{red}{rgb}{0.8, 0.2, 0.2}
\definecolor{darkred}{rgb}{0.6, 0.1, 0.05}
\definecolor{blueish}{rgb}{0.0, 0.3, .6}
\definecolor{light_gray}{rgb}{0.7, 0.7, .7}
\definecolor{pink}{rgb}{1, 0, 1}
\definecolor{greyblue}{rgb}{0.25, 0.25, 1}

\newcommand{\ApproachName}{DAE-Net\xspace}
\newcommand{\SupplementaryMaterial}{{{\color{dark_purple}Supplementary}}\xspace}

\DeclareMathOperator*{\argmax}{arg\,max}

\begin{document}
\title{\ApproachName{}: Deforming Auto-Encoder for fine-grained shape co-segmentation}

\author{Zhiqin Chen}
\affiliation{
 \institution{Adobe Research}
 \city{Seattle}
 \country{USA}
}
\email{zchen@adobe.com}

\author{Qimin Chen}
\affiliation{
 \institution{Simon Fraser University}
 \city{Burnaby}
 \country{Canada}
}
\email{qca43@sfu.ca}

\author{Hang Zhou}
\affiliation{
 \institution{Simon Fraser University}
 \city{Burnaby}
 \country{Canada}
}
\email{hza162@sfu.ca}

\author{Hao Zhang}
\affiliation{
 \institution{Simon Fraser University}
 \institution{Amazon}
 \city{Burnaby}
 \country{Canada}
}
\email{haoz@sfu.ca}

\begin{teaserfigure}
  \includegraphics[width=1.0\linewidth]{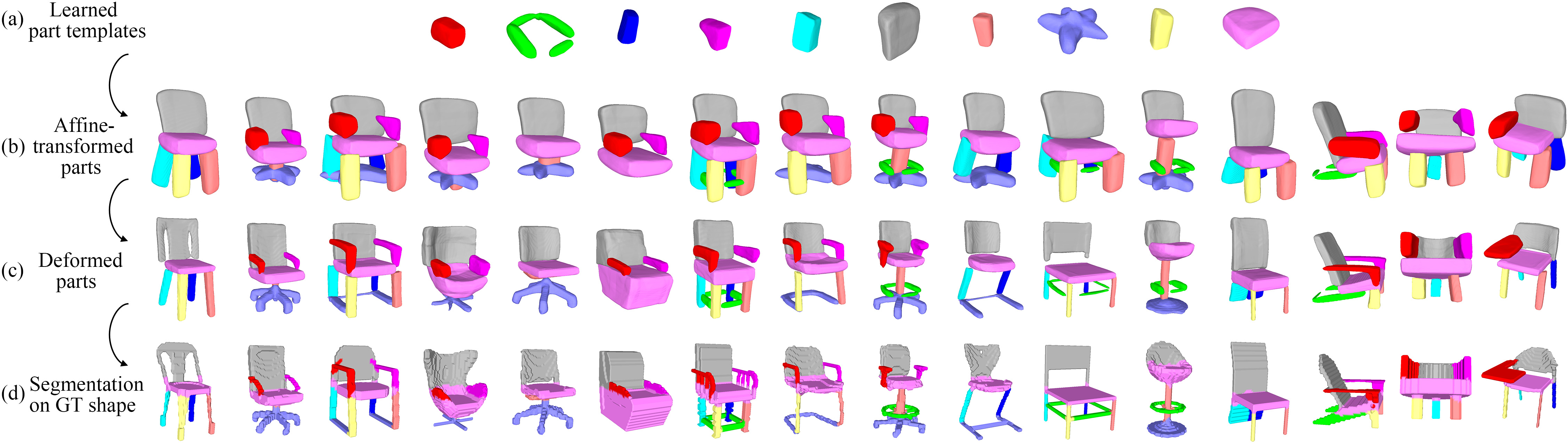}
  \caption{Our work \ApproachName{} co-segments a collection of 3D shapes into {\em fine-grained\/}, {\em consistent\/} parts. (a) The network learns a set of part templates shared by all shapes in the collection. (b) For each shape,  \ApproachName{} selects the required parts and assembles them via affine transforms. (c) Each transformed part is further refined through constrained deformation. (d) The deformed parts are finally mapped to the input shapes to obtain the shape co-segmentation. Our network is trained with a shape reconstruction loss and several regularization losses, without any part supervision. \vspace{6mm} }
  \label{fig:teaser}
\end{teaserfigure}

\newcommand\blfootnote[1]{%
  \begingroup
  \renewcommand\thefootnote{}\footnote{#1}%
  \addtocounter{footnote}{-1}%
  \endgroup
}

\begin{abstract}

We present an unsupervised 3D shape co-segmentation method which learns a set of {\em deformable part
templates\/} from a shape collection.
\blfootnote{ \textcopyright Authors | ACM 2024. This is the author's version of the work. It is posted here for your personal use. Not for redistribution. The definitive version of record was published in SIGGRAPH Conference Papers '24, https://doi.org/10.1145/3641519.3657528.}
To accommodate structural variations in the collection, our network composes each shape by a {\em selected subset\/} of template parts which are 
affine-transformed.
To maximize the expressive power of the part templates, we introduce a per-part deformation network to enable the modeling 
of diverse parts with substantial geometry variations, while imposing constraints on the deformation capacity to ensure fidelity to the originally 
represented parts.
We also propose a training scheme to effectively overcome local minima.
Architecturally, our network is a {\em branched autoencoder\/}, with a CNN encoder taking a voxel shape as input and producing per-part transformation
matrices, latent codes, and part existence scores, and the decoder outputting point occupancies to define the reconstruction loss.
Our network, coined DAE-Net for Deforming Auto-Encoder, can achieve unsupervised 3D shape co-segmentation that yields fine-grained, compact, and meaningful parts that are consistent across diverse shapes.
We conduct extensive experiments on the ShapeNet Part dataset, DFAUST, and an animal subset of Objaverse to show superior performance over prior methods.
Code and data are available at \url{https://github.com/czq142857/DAE-Net}.

\end{abstract}



\maketitle

\section{Introduction}
\label{sec:intro}

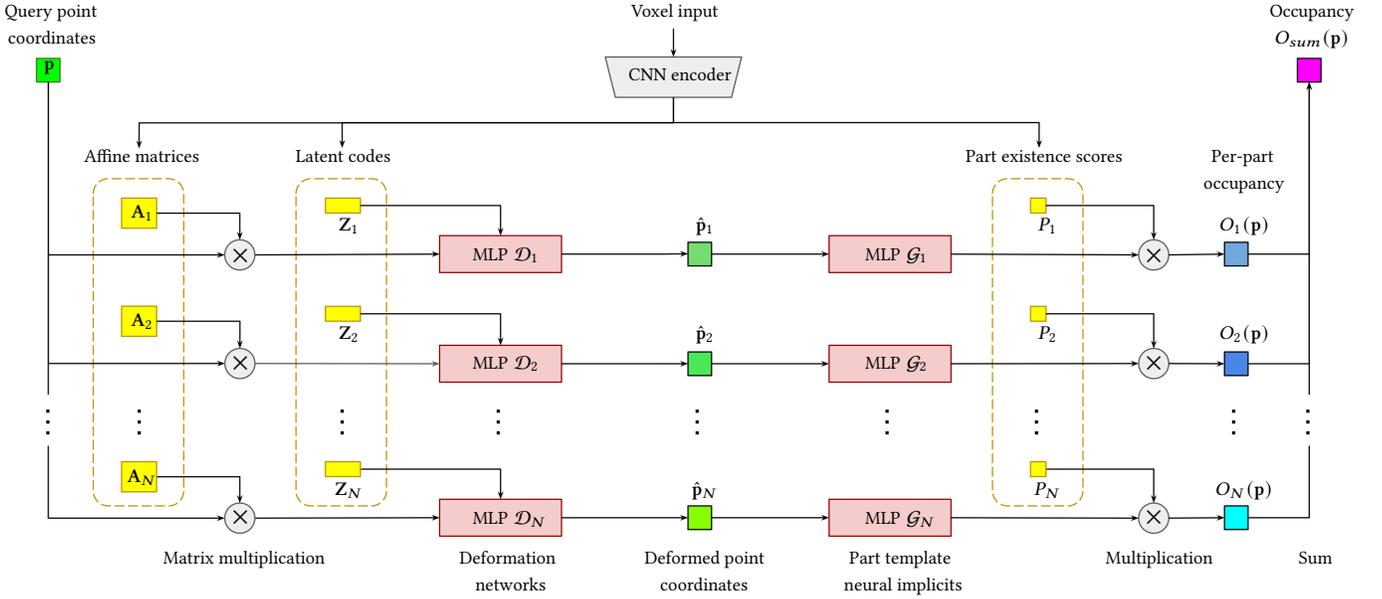
\begin{figure*}
\begin{picture}(522,220)
  \put(0,10){\includegraphics[width=1.0\linewidth]{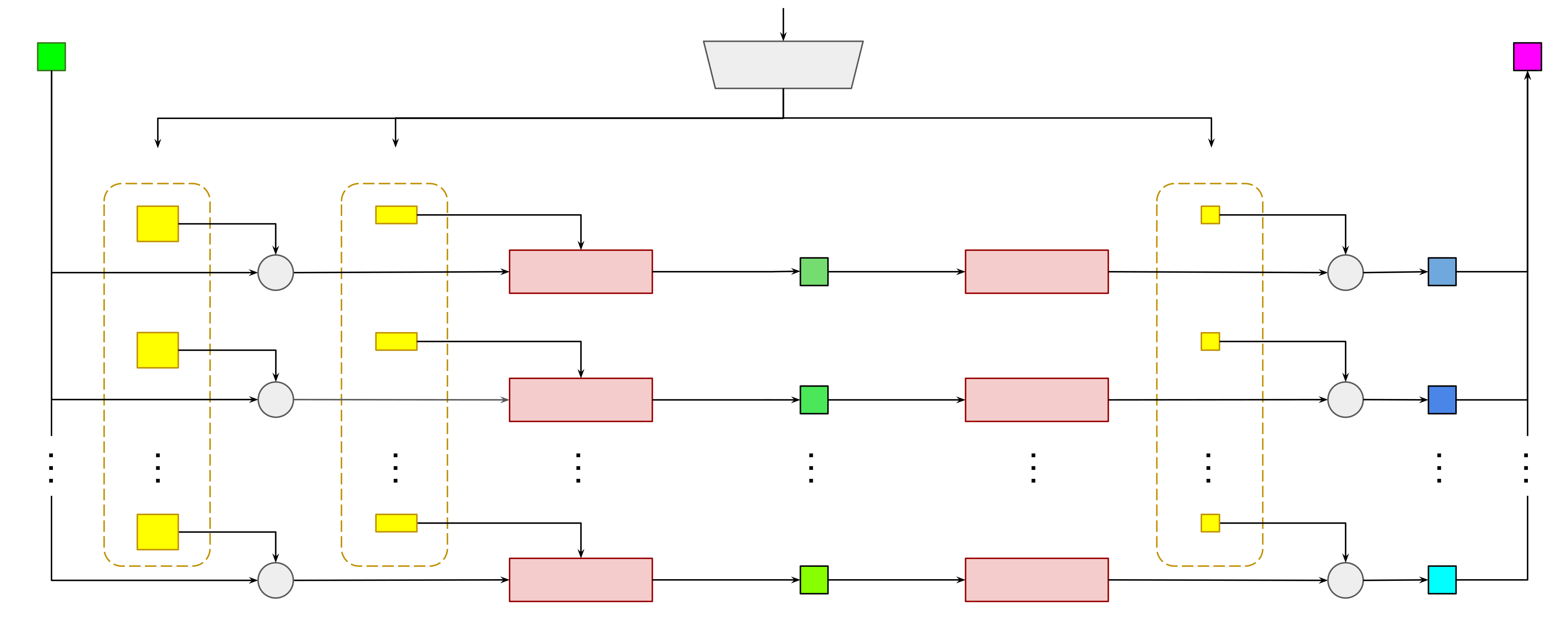}}
  \put(60, 13){\footnotesize Matrix multiplication}
  \put(172, 13){\footnotesize Deformation}
  \put(178, 3){\footnotesize networks}
  \put(242, 13){\footnotesize Deformed point}
  \put(248, 3){\footnotesize coordinates}
  \put(319.5, 13){\footnotesize Part template}
  \put(317.5, 3){\footnotesize neural implicits}
  \put(416.5, 13){\footnotesize Multiplication}
  \put(489.5, 13){\footnotesize Sum}

  \put(46.5, 44){\footnotesize $\mathbf{A}_{N}$}
  \put(125, 39){\footnotesize $\mathbf{Z}_{N}$}
  \put(86, 28){\large $\times$}
  \put(260, 39){\footnotesize $\hat{\mathbf{p}}_{N}$}
  \put(177, 28){\footnotesize MLP $\mathcal{D}_{N}$}
  \put(389.5, 39){\footnotesize $P_{N}$}
  \put(325.5, 28){\footnotesize MLP $\mathcal{G}_{N}$}
  \put(431.5, 28){\large $\times$}
  \put(458, 39){\footnotesize $O_{N}(\mathbf{p})$}

  \put(48, 103){\footnotesize $\mathbf{A}_{2}$}
  \put(126.5, 98){\footnotesize $\mathbf{Z}_{2}$}
  \put(86, 86.3){\large $\times$}
  \put(261.5, 98){\footnotesize $\hat{\mathbf{p}}_{2}$}
  \put(177, 87){\footnotesize MLP $\mathcal{D}_{2}$}
  \put(391, 98){\footnotesize $P_{2}$}
  \put(325.5, 87){\footnotesize MLP $\mathcal{G}_{2}$}
  \put(431.5, 86.3){\large $\times$}
  \put(459.5, 98){\footnotesize $O_{2}(\mathbf{p})$}

  \put(48, 144){\footnotesize $\mathbf{A}_{1}$}
  \put(126.5, 139){\footnotesize $\mathbf{Z}_{1}$}
  \put(86, 127){\large $\times$}
  \put(261.5, 139){\footnotesize $\hat{\mathbf{p}}_{1}$}
  \put(177, 128){\footnotesize MLP $\mathcal{D}_{1}$}
  \put(391, 139){\footnotesize $P_{1}$}
  \put(325.5, 128){\footnotesize MLP $\mathcal{G}_{1}$}
  \put(431.5, 127){\large $\times$}
  \put(459.5, 139){\footnotesize $O_{1}(\mathbf{p})$}

  \put(30, 165){\footnotesize Affine matrices}
  \put(110, 165){\footnotesize Latent codes}
  \put(363.5, 165){\footnotesize Part existence scores}
  \put(456.5, 165){\footnotesize Per-part}
  \put(453.5, 155){\footnotesize occupancy}

  \put(0, 220){\footnotesize Query point}
  \put(1.3, 210){\footnotesize coordinates}
  \put(15, 200){\footnotesize $\mathbf{p}$}
  \put(236, 196){\footnotesize CNN encoder}
  \put(237, 220){\footnotesize Voxel input}
  \put(478.5, 220){\footnotesize Occupancy}
  \put(480.5, 210){\footnotesize $O_{sum}(\mathbf{p})$}
\end{picture}
  \caption{
  \textbf{Network architecture of \ApproachName{}.}
Our network consists of $N$ branches representing $N$ parts of a 3D shape. To reconstruct part $i$, the query point coordinates in world frame are first transformed into the local frame of the part using an affine matrix that is predicted by a shape encoder network, a CNN. The transformed local coordinates are further deformed by a deformation MLP $\mathcal D_i$ conditioned on a latent code, that is both shape- and part-specific and also produced by the CNN, to refine the part details. Finally, the deformed local coordinates are fed into a part template MLP $\mathcal G_i$ to produce the occupancy of the query point. The occupancy is multiplied by the predicted part existence score from the CNN, so that the occupancy is set to zero if the part does not exist in the shape. We sum the occupancies from all $N$ parts to obtain the occupancy of the query point on the entire shape, which is used to compute the reconstruction loss.}
  \label{fig:network}
\end{figure*}

Co-analysis is a well-established paradigm for {\em unsupervised\/} learning over a data collection sharing some commonality, e.g., they all belong to the
same category~\cite{mitra_sigc13,xu_sigc16}. The central problem of co-analyzing a 3D shape collection is {\em shape co-segmentation\/}\footnote{It is worth distinguishing 3D shape co-segmentation
from object co-segmentation over an image collection, e.g.,~\cite{vicente2011ocs,chen2012vcs}. The latter is a special case of image segmentation with the goal of
jointly segmenting or {\em detecting\/} semantically similar {\em objects\/} out of multiple images or video frames.}, whose 
goal is to learn a consistent segmentation of all the shapes in the collection~\cite{Golov2009,sidi2011unsupervised,huang2013co,tulsiani2017learning,chen2019bae_net,zhu2020adacoseg,yang2021unsupervised,paschalidou2019superquadrics}.
On top of a structural understanding per shape, a co-segmentation induces part correspondences across the collection to facilitate a variety
of downstream tasks including attribute and knowledge transfer, shape editing or co-editing with part control, and generative modeling of shape
structures~\cite{gen3d_star}.

The most difficult challenge to shape co-segmentation arises when there are significant structural {\em and\/} geometric variations across the 
shape collection; see last row of Figure~\ref{fig:teaser}. Insisting on a single template for part organization, even a hierarchical one~\cite{vanKaick2013conshier}, would impose too 
much structural rigidity. As a result, the learned template is necessarily coarse to fulfill the consistency requirement across the whole collection.
On the flip side, abstraction-based methods using predefined primitives such as cuboids or superquadrics place rigidity on the modeling of part
geometries, often resulting in over-segmentation and less meaningful correspondences between the fragmented parts.

In this work, we present an unsupervised shape co-segmentation method which learns a set of {\em deformable part templates\/} from a collection of 3D shapes 
belonging to the same category. As shown in Figure~\ref{fig:teaser}, each learned part template (top row) models a set of corresponding parts in the shape collection. 
Our network is trained to transform, via affine transformations, and then deform a selected {\em subset\/} of the template parts to best reconstruct each shape in
the collection, without any part annotations as supervision.

The overall architecture of our network, as depicted in Figure~\ref{fig:network}, is that of an $N$-branch autoencoder, with $N$ being an upper bound on the number of
template parts for the shape category. A CNN encoder takes a voxelized shape as input and produces the affine matrices, part latent codes (to condition the
deforming MLPs), and part existence scores as a means to select among the part template MLPs to reconstruct the input shape. The part templates are modeled
as neural implicit functions, with the decoder outputting point occupancies to define the reconstruction loss.

Our network is coined {\em DAE-Net\/} for {\em Deforming Auto-Encoders\/}. 
The key idea behind our approach arises from the stipulation that corresponding parts in different shapes should have approximately the same 
shape (or form). This is inspired by the well-known design principle, ``form follows function,'' while part correspondence is ultimately about functional 
correspondence.
On the technical front, the design of our template learning has been inspired by Transforming Auto-encoders (TAE)~\cite{hinton2011transforming} 
for learning the first level of Capsule networks. Like TAE, our work represents a 3D shape via affine-transformed parts selected from a group of 
learned part templates, to accommodate structural variations and produce a fine-grained co-segmentation. However, the limited expressive power of affine transformations 
is not sufficient for faithfully representing parts with substantial geometric variations. Our per-part deformation alleviates such part modeling rigidity to learn diverse parts, 
while imposing constraints on the deformation capacity to ensure fidelity to the originally represented parts. 

In addition, we propose a training scheme to effectively overcome {\em local minima\/} encountered during training.
While BAE-Net, RIM-Net, and our method are all built on the branched auto-encoder structure and may suffer from local minima, only our method can adopt the proposed training scheme to overcome this issue. This is because the training scheme includes re-initializing certain auto-encoder branches, which is trivial in our method as it uses different networks to represent different parts (branches). However, it is non-trivial for the other two works since they produce all branches by the same MLP or require complex hierarchical training.

Through extensive experiments, we show that DAE-Net can achieve unsupervised 3D shape co-segmentation that yields fine-grained, compact, and meaningful 
parts that are consistent across diverse shapes. With comparisons conducted on the Shapenet Part dataset~\cite{chang2015shapenet,yi2016scalable}, DFAUST~\cite{dfaust}, and an animal subset of Objaverse~\cite{deitke2023objaverse},
DAE-Net exhibits superior performance over prior unsupervised shape segmentation methods.
Additionally, we demonstrate shape clustering using the part existence score produced by our method and showcase a controllable shape detailization application enabled by our segmentation results.

\section{Related work}
\label{sec:related}

\paragraph{3D co-segmentation with handcrafted priors.} 
Early works in geometry processing utilize fundamental geometric cues~\cite{shamir2008survey}, such as surface area, curvature, and geodesic distance, to derive higher-level semantic attributes for 3D shape segmentation. 
Golovinskiy and Funkhouser~\cite{golovinskiy2009consistent} investigated consistent co-segmentation of 3D shapes by constructing a graph connecting corresponding faces across different meshes. 
The unsupervised clustering approach was later extended to feature spaces via diffusion maps and spectral clustering~\cite{sidi2011unsupervised}. 
Thereafter, extensive research has been devoted to co-analysis of sets of shapes based on various clustering strategies~\cite{hu2012co, huang2011joint, huang2013co, meng2013unsupervised, xu2010style}. 
\cite{shu2016unsupervised} adapted the setting by transforming the handcrafted local features with stacked auto-encoders before per-shape graph cuts. 
In contrast, our approach is an end-to-end differentiable pipeline and free of externally introduced design elements previously proposed for shape segmentation.

\paragraph{Co-segmentation via 3D shape reconstruction.} 
To circumvent 3D annotations, weakly-supervised and unsupervised learning schemes utilize large collections of unlabelled data for segmentation. 
\cite{tulsiani2017learning} introduced sets of cuboids to approximate a 3D shape without supervision. 
\cite{sun2019learning} and \cite{yang2021unsupervised2} improved the cuboid representation to model complex shape structures.
\cite{paschalidou2019superquadrics} utilized superquadrics for shape abstractions, leading to better shape parsing ability. 
\cite{deprelle2019learning} represented shapes as deformation and combination of elementary structures represented in point clouds for shape reconstruction and unsupervised correspondence.
BAE-Net~\cite{chen2019bae_net} proposed a branched auto-encoder network for shape co-segmentation, where each individual branch learns to localize part instances across multiple samples.  
BSP-Net~\cite{chen2020bspnet} and CvxNet~\cite{deng2020cvxnet} represented shapes as convex polytopes using neural implicit fields, allowing shape co-segmentation in the view of commonly associated convexes. 
\cite{paschalidou2020learning} built a binary tree of primitives for shape reconstruction without part-level supervision. 
\cite{kawana2020neural} represented shapes as star primitives, where each primitive is formed by a continuous function defined on the sphere surface. 
Neural Parts~\cite{paschalidou2021neural} defined primitives as invertible neural networks, allowing inverse mapping between a sphere and the target part for efficient computation. 
RIM-Net~\cite{niu2022rim} represented shapes as hierarchical shape structures with recursive implicit fields.
PartNeRF~\cite{tertikas2023generating} represented objects as a collection of locally defined Neural Radiance Fields (NeRFs), and designed a part-aware generative model based on auto-encoders.
DPF-Net~\cite{shuai2023dpf} performed structured shape reconstruction by representing parts as deformed cuboids and cylinders.
\cite{huang2023learning} proposed to reconstruct part primitives from multi-view images while imposing convexity regularization on the parts.
Several works \cite{kim2023semantic,zheng2021deep,deng2021deformed} coupled neural implicit fields with neural deformation fields to learn dense correspondences between shapes.
Other than implicit representations, continuous progress of co-segmentation has been made for point clouds~\cite{zhu2020adacoseg, yang2022mil}.

Our work learns the {\em shapes\/} of a set of part templates. In contrast, prior works either had no explicit guidance for defining a part (e.g., BAE-Net, RIM-Net), merely relying on MLPs to perform segmentation, or make much stronger assumptions about the part templates, e.g., as cuboids, convexes, or other basic primitives. Our template represents the ``mean'' shape of the parts, and we use a subset of those templates to build each 3D shape via per-template affine transformation and local deformation. The learned templates also help achieve compactness of the segmentation, which is hard for primitive templates such as cuboids.

\paragraph{Zero-shot 3D segmentation using pretrained models.} 
Zero-shot segmentation aims to make predictions for categories that are not annotated in training. 
\cite{michele2021generative} extended zero-shot semantic image segmentation~\cite{bucher2019zero} to 3D, which is followed by \cite{chen2022zero, koo2022partglot}.
With the emergence of NeRFs~\cite{lombardi2019neural, mildenhall2021nerf}, methods have been developed to model semantic fields~\cite{zhi2021place, vora2021nesf, fan2022nerf, kundu2022panoptic, fu2022panoptic, tschernezki2022neural, siddiqui2023panoptic, hong20233d} by reconstructing semantic annotations from multi-view renderings. 

Going beyond zero-shot learning, the more general open-vocabulary setting~\cite{ding2023pla} assumes a large vocabulary corpus is accessible during training. Recently, great strides have been made in vision-language pretraining~\cite{alayrac2022flamingo, jia2021scaling, saharia2022photorealistic, zhang2022glipv2} by pretraining large-scale image-text pairs. 
Owing to learned rich visual concepts and notable zero-shot capabilities, 3D Highlighter~\cite{decatur20233d} and SATR~\cite{abdelreheem2023satr} proposed mesh segmentation based on off-the-shelf CLIP~\cite{radford2021learning} and GLIP~\cite{li2022grounded} models, and PartSLIP~\cite{liu2023partslip} relied on GLIP for point cloud segmentation.
\cite{abdelreheem2023zero} utilized large foundation vision and language models to perform co-segmentation between pairs of shapes.
By distilling zero-shot image segmentation models~\cite{li2022languagedriven, caron2021emerging} into NeRFs, \cite{kobayashi2022decomposing, goel2023interactive,peng2023openscene} are able to perform scene segmentation under open-vocabulary settings. In contrast, our method does not rely on any pretrained image-language models and is flexible on its own. 

\paragraph{Transforming Auto-encoders.}
\cite{hinton2011transforming} introduced Transforming Auto-encoders (TAE) to learn the first level of Capsule networks~\cite{sabour2017dynamic}. TAE learns to recognize visual entities, i.e., parts, and their poses, in an unsupervised manner by training an auto-encoder on a set of images. Specifically, each part $i$ is represented by a distinct capsule, which consists of a recognition module and a generation module. The recognition module predicts the probability $P_i$ that the part is present in input image $\mathcal{I}$, as well as the transformation of the part with respect to its canonical pose, e.g., translation $(x_i,y_i)$. The generation module $\mathcal{G}_i$ represents the ``shape'' of the part; it inputs the transformed coordinates and outputs the transformed part. The reconstructed image is obtained by 

\begin{align}
\mathcal{I}_{rec} = \sum_i P_i \cdot \mathcal{G}_i(x_i,y_i),
\end{align}

\noindent where the parts are selected by $P_i$ and transformed by $(x_i,y_i)$ to assemble the final output image.

\section{Method}
\label{sec:method}

In this section, we introduce the network architecture and loss functions of our method. We also present a training scheme to prevent the model from being stuck in a local minimum during optimization. Code and data are also provided in the \SupplementaryMaterial.

\subsection{Network architecture}

The network architecture of \ApproachName{} is shown in Figure~\ref{fig:network}. In our network, we use $N$ branches to present $N$ parts of the shape $\mathcal{V}$ to be reconstructed, where each part $i$ has a dedicated MLP $\mathcal{G}_i$ to represent the corresponding part template as a neural implicit~\cite{imnet,OccNet,DeepSDF}, and a dedicated deformation MLP $\mathcal{D}_i$ for part deformation. The shape encoder $\mathcal{E}$ is a 3D CNN that takes an occupancy voxel grid $\mathcal{V} \in \{0,1\}^{64 \times 64 \times 64}$ as input and produces per-part affine transformation matrices $\mathbf{A}_i^\mathcal{V} \in \mathbb{R}^{3 \times 4}$, latent codes $\mathbf{Z}_i^\mathcal{V} \in \mathbb{R}^{4}$, and part existence scores $P_i^\mathcal{V} \in [0,1]$, for part $i$. 

In the following, we assume that the outputs are all from shape $\mathcal{V}$ and therefore drop the superscript $^\mathcal{V}$ for simplicity. We also require densely sampled points from the shape $\mathcal{V}$ to train the neural implicit, and we denote the occupancy of a sampled point $\mathbf{p} \in \mathbb{R}^{3}$ as $\mathcal{V}[\mathbf{p}] \in \{0,1\}$.

To reconstruct part $i$, the query point $\mathbf{p}$ in world coordinates is first projected to homogeneous coordinates $\mathbf{p'} \in \mathbb{R}^{4}$, i.e., appending $1$ to $\mathbf{p}$, and then transformed into the local frame of the part via an affine transformation: $\mathbf{p}_i^{local} = \mathbf{A}_i \mathbf{p'}$.
The local coordinates $\mathbf{p}_i^{local}$ are further deformed by the deformation MLP $\mathcal{D}_i$ conditioned on the latent code $\mathbf{Z}_i$. Note that $\mathcal{D}_i$ predicts the offsets of the deformation: $\Delta \mathbf{p}_i^{local} = \mathcal{D}_i(\mathbf{p}_i^{local}, \mathbf{Z}_i)$, therefore the deformed coordinates are $\hat{\mathbf{p}}_i = \mathbf{p}_i^{local} + \Delta \mathbf{p}_i^{local}$.

Finally, the deformed coordinates $\hat{\mathbf{p}}_i$ are fed into the part template MLP $\mathcal{G}_i$ to produce the occupancy $\mathcal{G}_i(\hat{\mathbf{p}}_i) \in [0,1]$.
The occupancy is multiplied by the predicted part existence score $P_i$, so that the occupancy will be set to zero if the part is not deemed to exist in the shape. The final occupancy for point $\mathbf{p}$ on part $i$ is $O_i(\mathbf{p}) = P_i \cdot \mathcal{G}_i(\hat{\mathbf{p}}_i)$.

To obtain occupancy of the query point on the entire shape, we sum the occupancies from all parts: $O_{sum}(\mathbf{p}) = \sum_i O_i(\mathbf{p})$. Since the parts should ideally be non-overlapping, i.e., the per-part occupancies should be one-hot, $\sum_i O_i(\mathbf{p}) = \max_i O_i(\mathbf{p}) = 1$, for occupied points, we also use an auxiliary shape occupancy $O_{max}(\mathbf{p}) = \max_i O_i(\mathbf{p})$ to compute the reconstruction losses.

\subsection{Loss functions}

First, we use a shape reconstruction loss,

\begin{align}
\mathcal{L}_{recon}^{sum} = \mathbb{E}_{\mathbf{p}} ( O_{sum}(\mathbf{p})-\mathcal{V}[\mathbf{p}] )^2.
\end{align}

to supervise the entire model.  As mentioned, to encourage non-overlapping parts, we use a secondary reconstruction loss,

\begin{align}
\mathcal{L}_{recon}^{max} = \mathbb{E}_{\mathbf{p}} ( O_{max}(\mathbf{p})-\mathcal{V}[\mathbf{p}] )^2.
\end{align}

Second, we need to constrain the per-part deformation, so that the deformation only changes the part locally and does not transform it into a different part or multiple parts. We use a simple $L_2$ loss on the predicted deformation offsets to achieve this

\begin{align}
\mathcal{L}_{deform} = \mathbb{E}_{\mathbf{p}} \mathbb{E}_{i} ||\Delta \mathbf{p}_i^{local}||_2^2.
\end{align}

Finally, we have a loss to control the sparsity of the segmented parts, which is achieved by penalizing the predicted part existence scores. Specifically, in each training iteration, for a mini-batch of shapes $\mathcal{S}$, we have

\begin{align}
\mathcal{L}_{sparse} = - \mathbb{E}_i ( 1 - \mathbb{E}_{\mathcal{V} \in \mathcal{S}} P_i^\mathcal{V} )^2,
\end{align}

where $\mathbb{E}_{\mathcal{V} \in \mathcal{S}} P_i^\mathcal{V}$ evaluates what fraction of the shapes in $\mathcal{S}$ contains part $i$. If part $i$ is rare, $\mathbb{E}_{\mathcal{V} \in \mathcal{S}} P_i^\mathcal{V}$ is close to zero and $P_i$ will receive a greater penalty, therefore effectively making uncommon parts disappear.

The overall loss is a weighted sum of the loss terms,

\begin{align}
\mathcal{L} = \mathcal{L}_{recon}^{sum} + \alpha \mathcal{L}_{recon}^{max} + \beta \mathcal{L}_{deform} + \gamma \mathcal{L}_{sparse}.
\end{align}

We set $\alpha = 0.1$ and $\beta=100$. $\gamma$ needs to be set according to the desired granularity of the segmentation; see Figure~\ref{fig:visual_ablation_sparsity}.
Typically, different shape categories require different $\gamma$ values to achieve desirable segmentation quality, therefore tuning $\gamma$ is often necessary.
In our experiments, we try $\gamma=0.02, 0.01, 0.002,$ and $0.001$ in decreasing order until reaching the desired granularity.

\begin{table*}[t!]
\caption{
\textbf{Quantitative results on shape segmentation} compared to BAE-Net~\cite{chen2019bae_net} and RIM-Net~\cite{niu2022rim}, evaluated by average per-part IOU. Best results are marked in bold. \vspace{-4mm}
}
\label{table:comparison}
\begin{center}
\resizebox{\linewidth}{!}{
\begin{tabular}{l|r |r|r|r|r|r |r|r|r|r|r  |r|r|r|r|r }
\hline
 & \textbf{Mean} & plane & bag & cap & chair & earph. & guitar & knife & lamp & laptop & motor. & mug & pistol & rocket & skateb. & table \\
\hline\hline
BAE-Net & 56.2 & 59.8 & 84.4 & 84.9 & 54.1 & 44.1 & 51.0 & 32.5 & {\bf 74.7} & 27.1 & 27.5 & {\bf 94.4} & 29.0 & {\bf 40.9} & 63.3 & 75.7\\
RIM-Net & 53.6 & 52.7 & {\bf 86.1} & 62.6 & 79.2 & 72.9 & 25.7 & 29.5 & 68.3 & 33.2 & 28.5 & 48.6 & 36.2 & 39.5 & 64.9 & {\bf 76.0}\\
Ours & {\bf 76.9} & {\bf 78.0} & 84.4 & {\bf 86.3} & {\bf 85.5} & {\bf 77.2} & {\bf 88.4} & {\bf 85.8} & 73.2 & {\bf 95.0} & {\bf 48.1} & 94.2 & {\bf 74.6} & 38.7 & {\bf 68.2} & 75.5\\
\hline
\end{tabular}
}
\end{center}
\end{table*}
\begin{figure*}[t!]
\begin{center}
\includegraphics[width=1.0\linewidth]{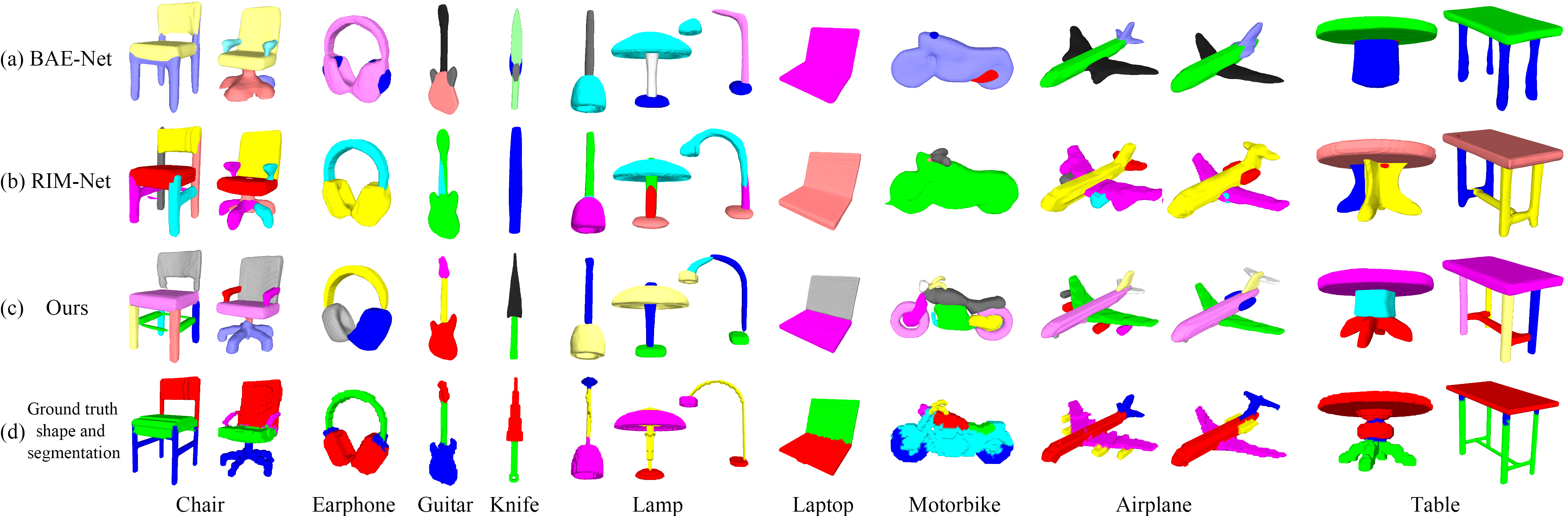}
\end{center}
\caption{
\textbf{Qualitative results on shape segmentation} compared to BAE-Net~\cite{chen2019bae_net} and RIM-Net~\cite{niu2022rim} on ShapeNet Part dataset~\cite{chang2015shapenet,yi2016scalable}. Within the same category, same color indicates the parts are from the same branch of the network, thus are considered to be corresponded.
Since the ground truth segmentation in ShapeNet Part dataset is on point clouds, we color the voxels in (d) using nearest neighbor.
}
\label{fig:visual_comparison}
\end{figure*}

\subsection{Overcoming local minima}

Since unsupervised co-segmentation via reconstruction is an ill-posed problem, the training is often stuck in a local minimum, where a predicted part represents multiple ground truth parts, or a number of predicted parts represent a single ground truth part, and further training cannot break the part apart or merge the parts together; see Figure~\ref{fig:visual_ablation_network} (a).

To devise a training scheme to overcome local minima, we partition the training into $K$ eras, each containing $M$ iterations. Then we design an operation \textit{revive($i$)}, to re-initialize the weights of $\mathcal{D}_i$, $\mathcal{G}_i$, and the portion of the last layer of $\mathcal{E}$ that affects $\mathbf{A}_i$, $\mathbf{Z}_i$, and $P_i$.
In the first era, we initialize $\mathcal{E}$ and revive all the branches. We explicitly track the age of each branch, i.e., how many eras a specific branch has lived after its last revival. At the end of each era, we compute $\mathbb{E}_{\mathcal{V} \in \mathcal{S}} P_i^\mathcal{V}$ for all the training shapes $\mathcal{S}$, which indicates what fraction of the training shapes contains part $i$. If the fraction for part $i$ is lower than a threshold (10\%), the branch is considered ``dead'' and will be revived. We also revive the oldest branch, so the part it represents has the option of breaking down into smaller parts or being merged into other parts. Reviving the oldest branch holds the risk that it can make the model stuck in a worse local minimum. Therefore, at the end of each era, we also compute the average reconstruction error evaluated by IoU (Intersection over Union) of per-point occupancy, and compare it with the IoU of the previous era. If the current IoU is better than the previous one, we keep the current network weights; otherwise we load the network weights of the previous era. This is based on the assumption that a better segmentation should always lower the reconstruction error.

\subsection{Training details}
In the experiments, we train individual models for different object categories. We set branch number $N=16$. However, for categories with evidently less parts, such as laptops and mugs, we set $N=8$ to save training time. We train each model $2N$ eras with each era consisting of $125,000$ iterations of training. In each iteration, we set the mini-batch size to $16$. The model is trained with Adam optimizer~\cite{kingma2014adam} with a learning rate of $0.0002$. Training on one category takes $8$ hours when $N=16$ and $2.5$ hours when $N=8$, on one NVIDIA V100 GPU.

\section{Experiments}
\label{sec:exps}

Our experiments mainly focus on unsupervised co-segmentation.
Since our method predicts an existence score for each shape part, we can group the shapes in the dataset according to which parts they have.
Finally, we show that our segmentation results can be easily combined with DECOR-GAN~\cite{chen2021decor} to achieve shape detailization while controlling per-part geometric style.

\subsection{Unsupervised shape co-segmentation}

We perform experiments on shapes from 15 categories in the ShapeNet Part dataset ~\cite{chang2015shapenet,yi2016scalable}; the car category is excluded since its ground truth segmentation consists of non-volumetric parts; see Section~\ref{sec:conclusions} and Figure~\ref{fig:failure} (a) for details. We compare with BAE-Net~\cite{chen2019bae_net} and RIM-Net~\cite{niu2022rim}, which also perform unsupervised shape co-segmentation. There are other works that perform shape abstraction thus can potentially be used for shape segmentation, but they either over-segment the shape into excessive numbers of parts \cite{chen2020bspnet,deng2020cvxnet}, or do not show consistent part correspondence across shapes \cite{paschalidou2019superquadrics,deprelle2019learning,paschalidou2021neural}, therefore they are not being compared in the experiments. For all methods, we train an individual model on all shapes in each shape category. All methods are trained on pre-processed voxel data provided by \cite{chen2019bae_net}. We use the default settings of BAE-Net and RIM-Net to train their models, except that their original branch numbers are $8$, therefore we try both $8$ and $16$ as their branch numbers for each category and show the best results.

We quantitatively evaluate the co-segmentation accuracy by per-part IOU averaged on all parts and all shapes in the category, which is the standard evaluation metric used for shape segmentation on the ShapeNet Part dataset. To infer segmentation from the network, for each query point $p$, its label will be assigned as the label of the network branch that produces the maximum occupancy, i.e., $label(p) = label(\argmax_i O_i(\mathbf{p}))$. To achieve objectiveness and fairness in evaluation, for all methods, we use an algorithm to automatically label each output branch of the networks with a semantic part label so that the average IOU is maximized. The automatic labeling algorithm is implemented by exhaustive search. The average IOU is computed on the standard testing split of the ShapeNet Part dataset for each category, so that the reported IOU can be directly compared with results from other semi-supervised or fully-supervised methods on this dataset.

As shown by the quantitative results from Table~\ref{table:comparison}, our method outperforms the compared methods by a large margin on most categories. On the few categories where the other methods performed better, our method is often less than $2\%$ worse than the best-performing alternative.

Importantly however, these quantitative results do not tell the whole story, since the ground truth segmentation is coarse, e.g., chair is only segmented into 4 high-level parts: back, seat, leg, and arm. DAE-Net has been designed to provide {\em fine-grained\/} segmentation, as shown in Figure~\ref{fig:visual_comparison}, while other methods either fail to segment finer parts or have messy segmentations. We also show the learned part templates in the \SupplementaryMaterial.

\begin{figure}[t!]
\begin{center}
\includegraphics[width=1.0\linewidth]{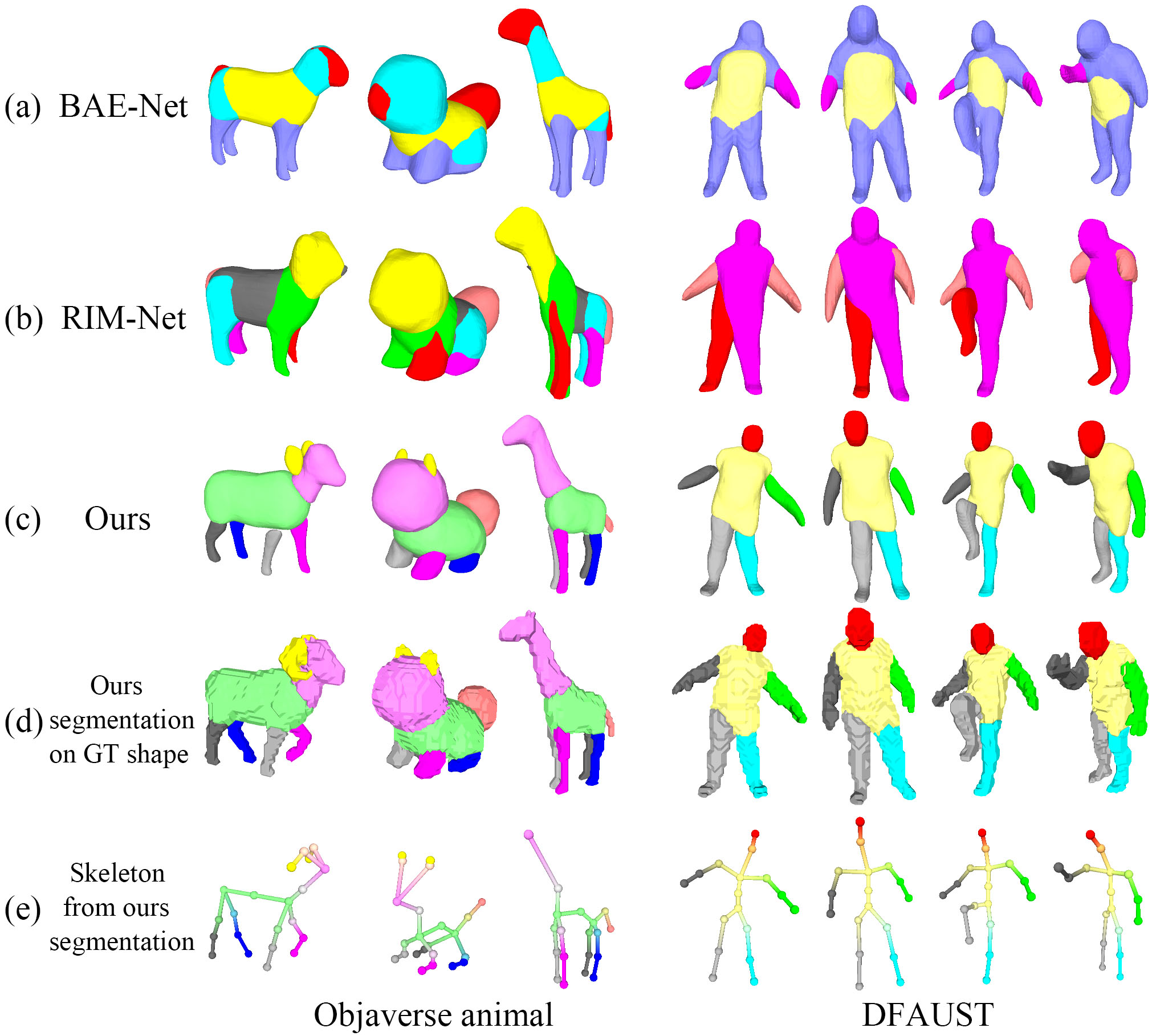}
\end{center}
\caption{
\textbf{Qualitative results on shape segmentation} on an animal subset of Objaverse~\cite{deitke2023objaverse}, and DFAUST~\cite{dfaust}. We also show the skeletons built upon our segmentation. More results can be found in Figure~\ref{fig:supp_animal_2x} and the \SupplementaryMaterial.
}
\label{fig:visual_comparison_extra}
\end{figure}

\begin{figure*}[t!]
\begin{center}
\includegraphics[width=1.0\linewidth]{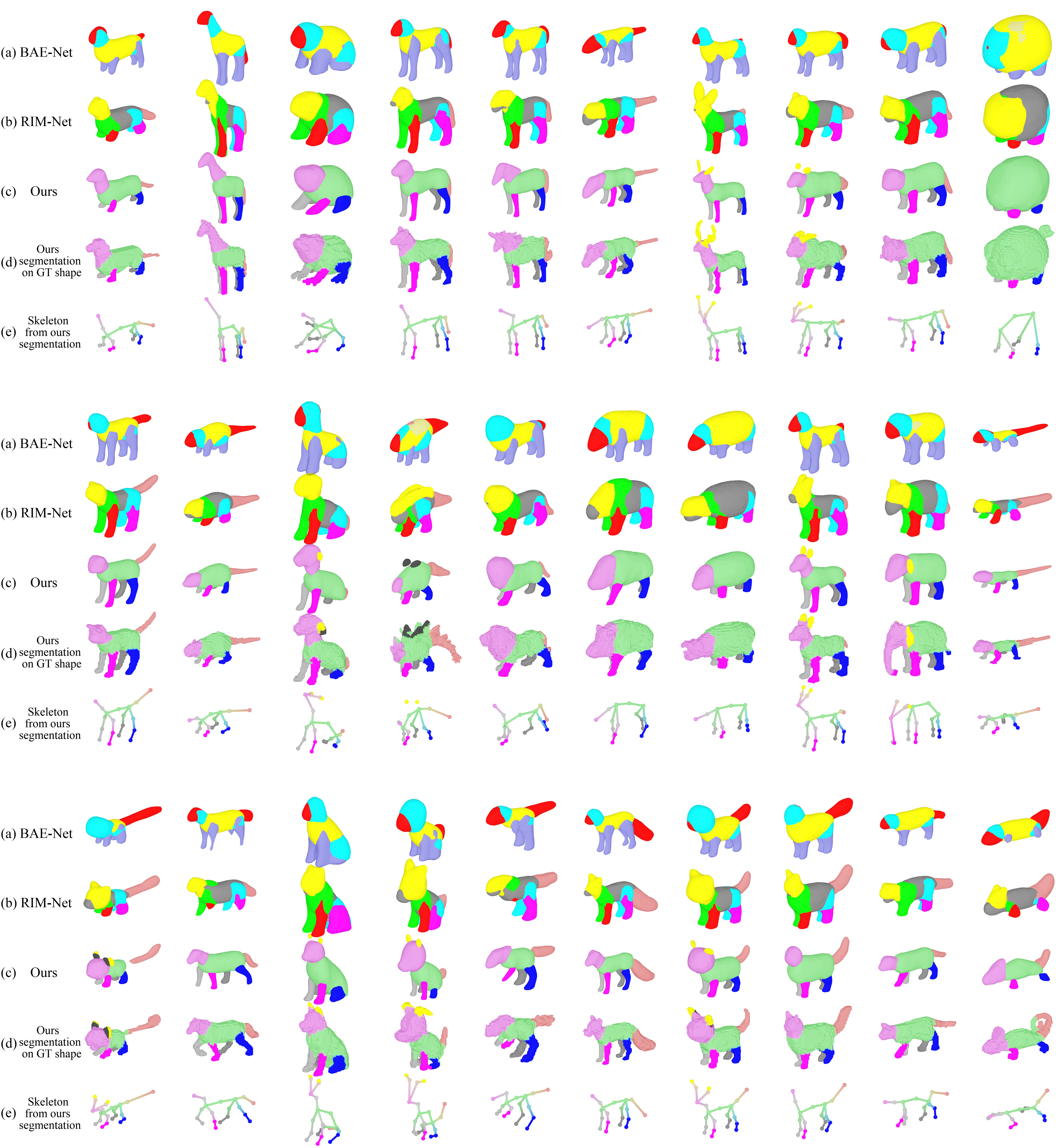}
\end{center}
\caption{
\textbf{Qualitative results on shape segmentation} on the \textbf{quadruped animal subset of Objaverse}, compared with other methods. Within the same category, same color indicates the parts are from the same branch of the network, thus are considered to be corresponded. Notice how semantically consistent our segmented parts are, when compared with other methods. We also show the skeletons built upon our segmentation.
}
\label{fig:supp_animal_2x}
\end{figure*}

In addition, we trained all methods on two additional datasets: DFAUST~\cite{dfaust} that contains human body shapes, and a subset of Objaverse~\cite{deitke2023objaverse} that contains quadruped animals. Qualitative results are shown in Figure~\ref{fig:visual_comparison_extra} and Figure~\ref{fig:supp_animal_2x}, where our method has clearly better performance. Moreover, since our method produces clean segmentation with part correspondences between shapes, we can easily build a skeleton of the shape using our segmentation results, as shown in Figure~\ref{fig:visual_comparison_extra} (e) and Figure~\ref{fig:supp_animal_2x} (e).
Since DFAUST and Objaverse do not have ground truth segmentation, quantitative results are hard to obtain. Therefore, we provide more details and ample qualitative results in the \SupplementaryMaterial.

\begin{figure}[t!]
\begin{center}
\includegraphics[width=1.0\linewidth]{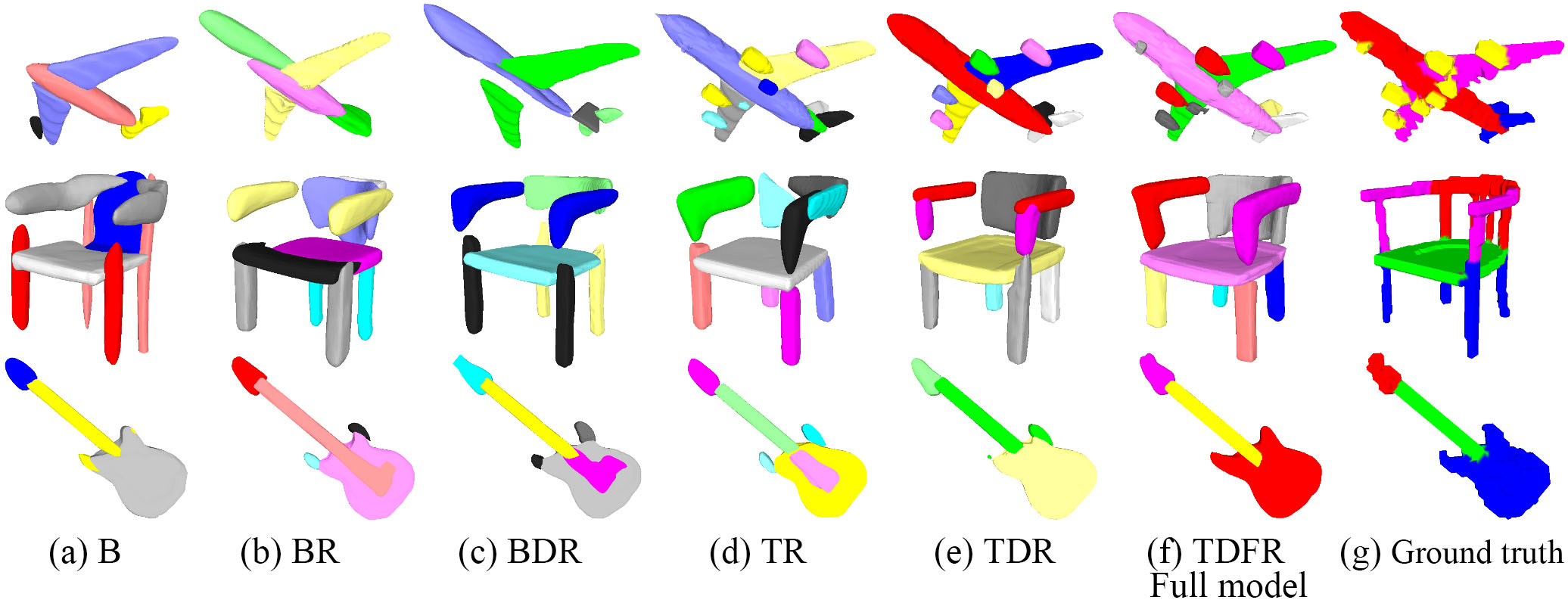}
\end{center}
\caption{
\textbf{Qualitative results of Ablation study} on airplane, chair, and guitar.
See Section~\ref{subsec:ablation} for the meaning of the abbreviations.
}
\label{fig:visual_ablation_network}
\end{figure}

\begin{figure}[t!]
\begin{center}
\includegraphics[width=1.0\linewidth]{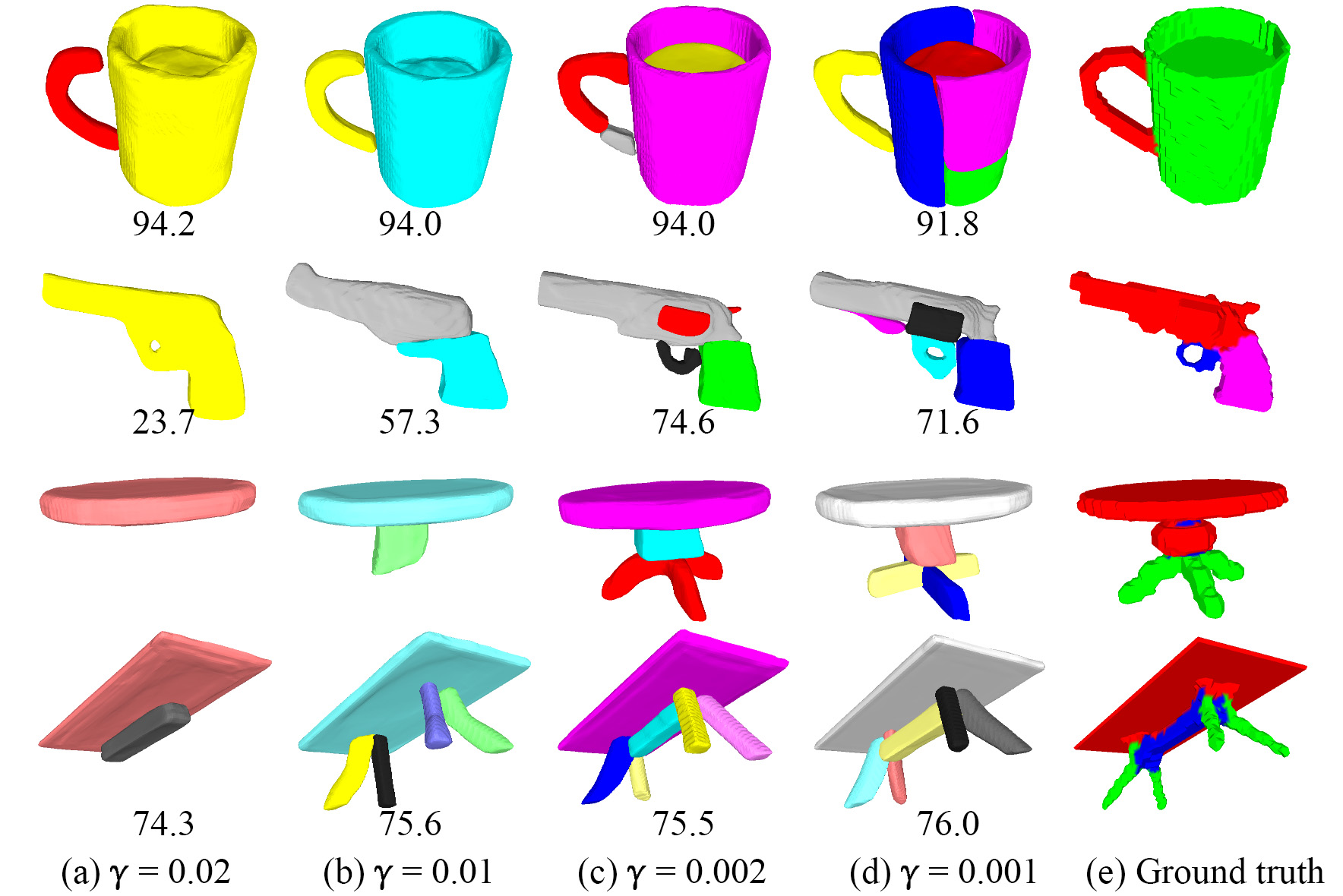}
\end{center}
\caption{
\textbf{Ablation study on the weight $\gamma$ of the sparsity loss $\mathcal{L}_{sparse}$} on mug, pistol, and table. The number under each shape shows the IOU of its category when trained with a specific $\gamma$ value. 
}
\label{fig:visual_ablation_sparsity}
\end{figure}

\begin{table*}[t!]
\caption{
\textbf{Ablation study} on shape segmentation, evaluated by average per-part IOU. See Section~\ref{subsec:ablation} for the meaning of the abbreviations.
Mean is the mean IOU on all 15 categories. We also report some IOUs on representative categories.
}
\label{table:ablation}
\begin{center}
\resizebox{\linewidth}{!}{
\begin{tabular}{l ||c|c|c|c |c|c|c|c ||c|c|c|c |c|c|c|c }
\hline
 & B$_\textrm{S}$ & B$_\textrm{M}$ & B & BD & BDF & T & TD & TDF & B$_\textrm{S}$R & B$_\textrm{M}$R & BR & BDR & BDFR & TR & TDR & TDFR (Full model)\\
\hline\hline
\textbf{Mean} & 59.1 & 45.1 & 58.6 & 62.3 & 57.8 & 61.5 & 68.9 & 68.6 & 69.3 & 60.6 & 73.2 & 72.6 & 70.3 & 74.6 & 74.5 & {\bf 76.9}\\
Plane & 56.2 & 30.5 & 53.3 & 71.2 & 62.6 & 60.8 & 65.7 & 73.6 & 71.8 & 53.8 & 71.3 & 74.4 & 72.5 & 75.1 & {\bf 78.0} & {\bf 78.0}\\
Chair & 73.3 & 49.2 & 60.5 & 69.9 & 73.4 & 69.3 & 74.9 & 76.0 & 84.7 & 57.6 & 84.5 & 83.5 & 84.0 & 85.2 & 84.2 & {\bf 85.5}\\
Guitar & 52.9 & 27.5 & 78.4 & 46.4 & 53.6 & 70.2 & 84.0 & 73.2 & 77.1 & 37.1 & 85.5 & 86.6 & 86.4 & 88.1 & 84.1 & {\bf 88.4}\\
\hline
\end{tabular}
}
\end{center}
\end{table*}
\begin{table*}[t!]
\caption{
\textbf{Quantitative results on few-shot shape classification}, and number of shapes for each category in ShapeNet Part dataset.
}
\label{table:classification}
\begin{center}
\resizebox{\linewidth}{!}{
\begin{tabular}{l |r|r|r|r|r |r|r|r|r|r  |r|r|r|r|r }
\hline
 & plane & bag & cap & chair & earph. & guitar & knife & lamp & laptop & motor. & mug & pistol & rocket & skateb. & table \\
\hline\hline
Classification precision & 0.93 & 0.79 & N/A & 0.97 & N/A & 0.96 & 0.83 & 0.86 & 0.96 & 0.84 & 0.76 & 0.87 & 0.44 & 0.30 & 0.91 \\
Classification recall & 0.98 & 0.14 & 0.00 & 0.96 & 0.00 & 0.99 & 0.94 & 0.73 & 1.00 & 0.86 & 0.90 & 0.85 & 0.73 & 0.07 & 0.91 \\
Number of shapes & 2,690 & 76 & 55 & 3,746 & 69 & 787 & 392 & 1,546 & 445 & 202 & 184 & 275 & 66 & 152 & 5,263 \\
\hline
\end{tabular}
}
\end{center}
\end{table*}

\subsection{Ablation studies}
\label{subsec:ablation}

Our full model is made of several components. We start from the base model (\textbf{B}) where we have $N$ MLP branches $\mathcal{G}_i$, each taking point coordinates $p$ and shape latent code as input and outputting the occupancy of $p$ in part $i$; the per-part occupancies are weighted by the predicted part existence scores $P_i$ to produce the final output.
In the base model \textbf{B}. we use both $\mathcal{L}_{recon}^{sum}$ and $\mathcal{L}_{recon}^{max}$ in the loss function.
We could use only $\mathcal{L}_{recon}^{sum}$ for training, denoted as \textbf{B$_\textrm{S}$}; or only $\mathcal{L}_{recon}^{max}$, denoted as \textbf{B$_\textrm{M}$}.
Next, we introduce transforming auto-encoder, denoted as \textbf{T}, where $\mathcal{G}_i$ is no longer conditioned on latent codes since it is now a part template shared by all shapes; and the templates will be affine-transformed by the predicted matrices.
For both \textbf{B} and \textbf{T}, we further add the deformation networks $\mathcal{D}_i$ to deform each individual part, denoted as \textbf{D}. We include $\mathcal{L}_{deform}$ to constrain the deformation, denoted as \textbf{F}. Finally, we apply our training scheme to overcome the local minima, denoted as \textbf{R}.

Table~\ref{table:ablation} summarizes the quantitative results, where we report the mean IOU on all categories and some IOUs on representative categories. We also show some qualitative results in Figure~\ref{fig:visual_ablation_network}. Our training scheme R plays a major role in improving the results. Methods trained without R tend to be stuck in local minima and cannot be improved further. B$_\textrm{M}$ and B$_\textrm{M}$R perform much worse than B$_\textrm{S}$ and B$_\textrm{S}$R, showing that $\mathcal{L}_{recon}^{max}$ is not as easy to optimize as $\mathcal{L}_{recon}^{sum}$ since its gradient cannot be propagated to all network branches. However, $\mathcal{L}_{recon}^{max}$ can help some categories achieve better results, as shown by B$_\textrm{S}$R vs. BR. Transforming auto-encoder with our training scheme (TR) can already achieve impressive results, but after augmented with deforming networks (D) with constraints (F), our full model (TDFR) performs the best.

Note that the sparsity loss $\mathcal{L}_{sparse}$ is applied to all the cases above. Its weight $\gamma$ in the final objective function controls the granularity of the segmentation results. We vary $\gamma$ for some shape categories and show quantitative and qualitative results in Figure~\ref{fig:visual_ablation_sparsity}.

\subsection{Shape clustering}

\begin{figure*}[t!]
\begin{center}
\includegraphics[width=1.0\linewidth]{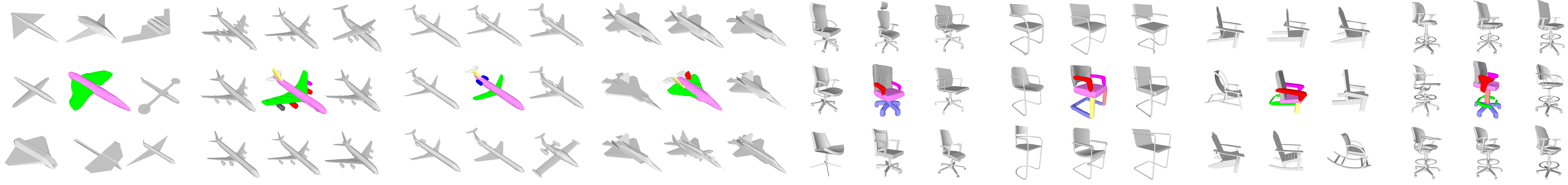}
\end{center}
\caption{
\textbf{Shape clustering} using the part existence score. Each 3x3 sub-figure shows a clustered group. The center of each sub-figure shows a reconstructed shape representing the parts that shapes in this group should have; the rest shows the first few shapes in the group.
}
\label{fig:grouping}
\end{figure*}

After thresholding the predicted part existence scores $P_i$ with a pre-defined threshold ($0.5$), we obtain a binary $N$-d vector representing whether part $i$ exists in the output shape or not. We can then group the shapes into sub-categories according to this vector. In Figure~\ref{fig:grouping}, we show some groups for airplane and chair. The clustering produces groups that contain structurally similar shapes.

\begin{figure}[t!]
\begin{center}
\includegraphics[width=1.0\linewidth]{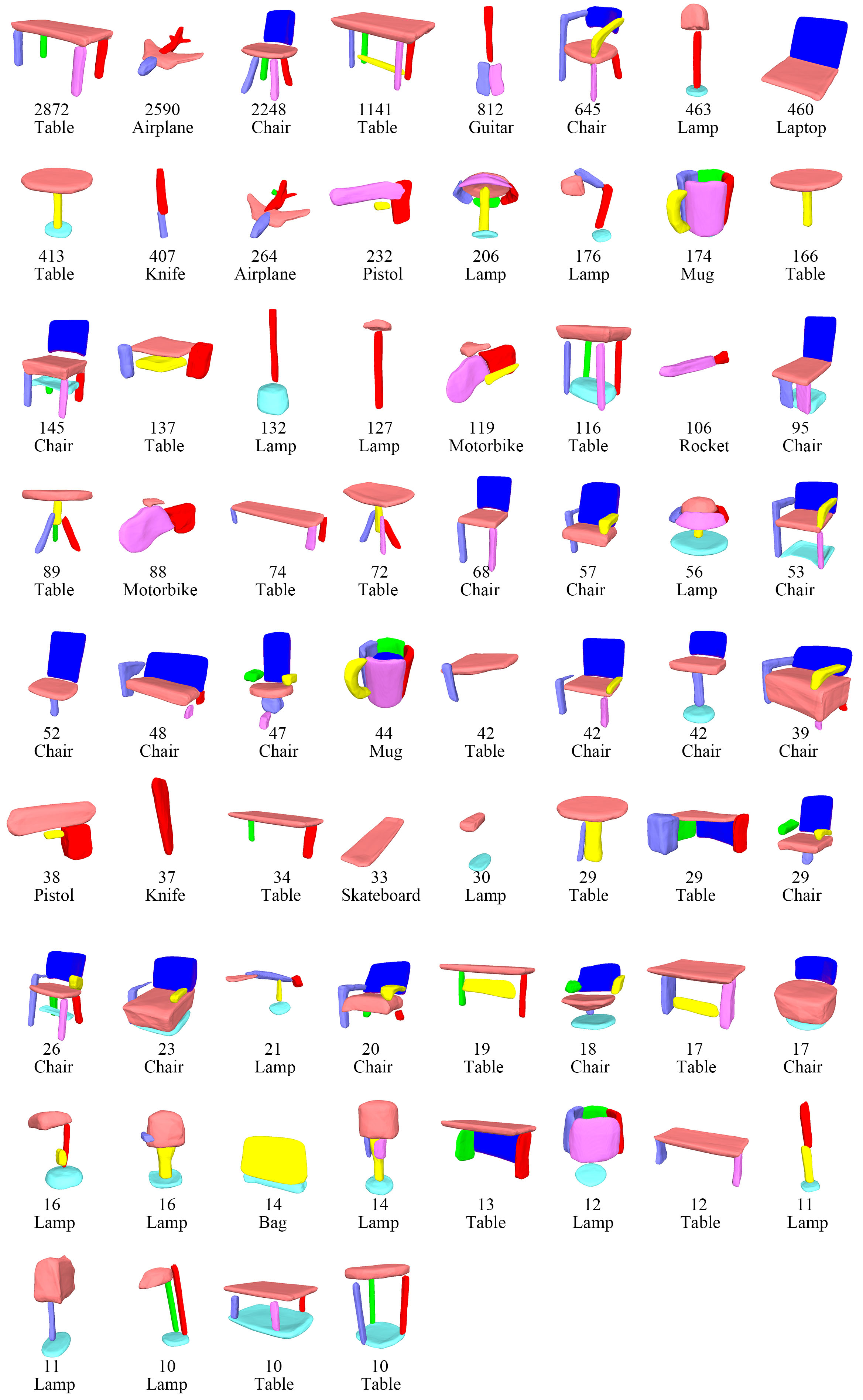}
\end{center}
\caption{
\textbf{Shape clusters} when trained on all the categories in ShapeNet Part dataset. We show all groups that contain at least 10 shapes. For each group, we show a representative reconstructed shape, the number of shapes it contains, and its semantic label for computing the numbers in Table~\ref{table:classification}.
}
\label{fig:grouping_allcat}
\end{figure}

We additionally train a single model on all categories in the ShapeNet Part dataset, which produces $68$ groups that contain no less than 10 shapes. We label each group with the closest category label, as shown in Figure~\ref{fig:grouping_allcat}. For other groups with less than 10 shapes, we assign to them an ``N/A'' label. Therefore, we have effectively classified all the shapes into different categories. We compare our classification results with the ground truth to obtain the quantitative results provided in Table~\ref{table:classification}. Our ``few-shot'' classification model has achieved decent results on categories with abundant shapes but ignored rare categories.

\subsection{Application: part-level shape detailization}

\begin{figure}[t!]
\begin{center}
\includegraphics[width=1.0\linewidth]{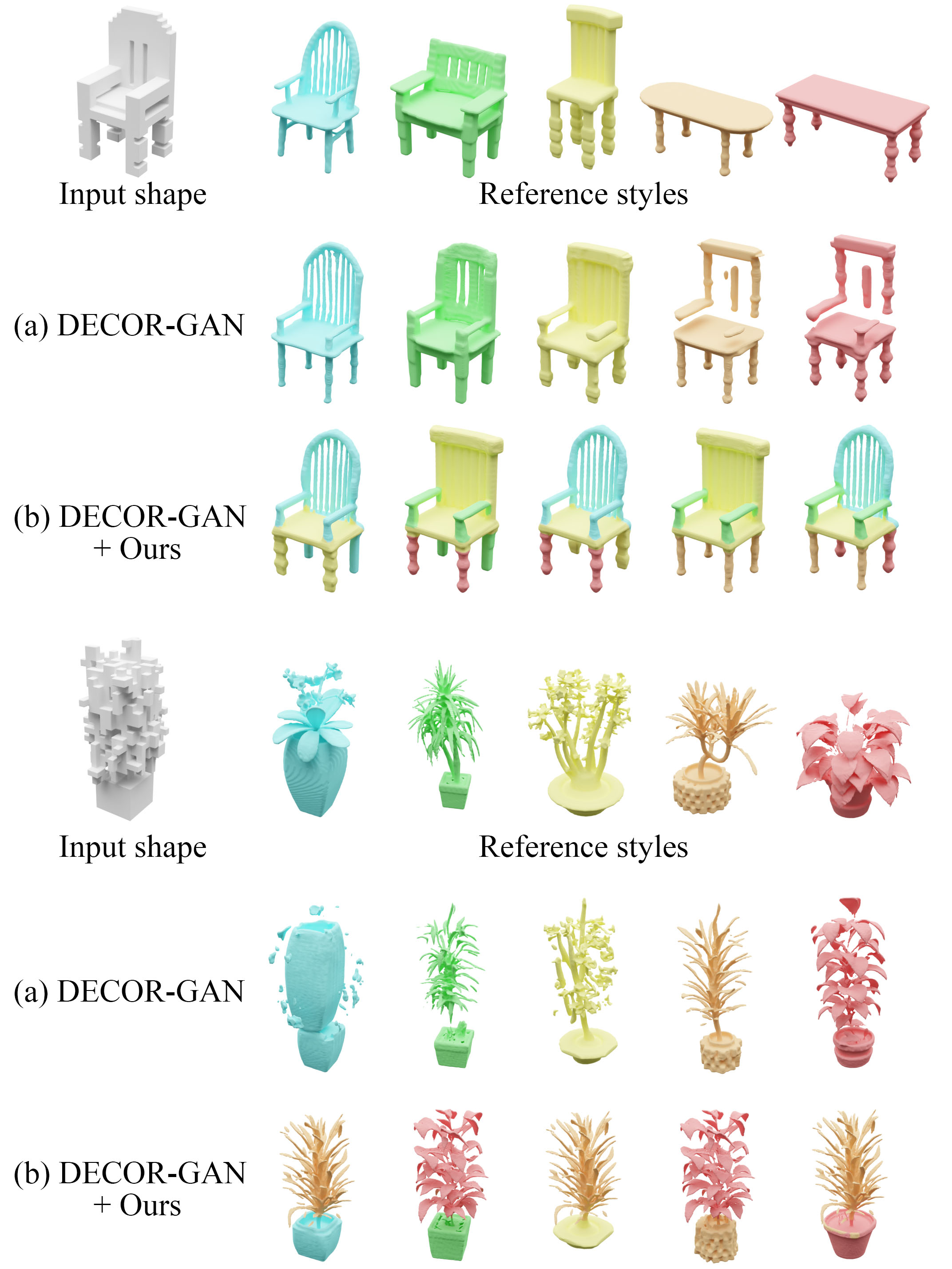}
\end{center}
\caption{
\textbf{Part-level shape detailization} on chair-table and plant. We use different colors to distinguish different geometric styles. To generate geometric details on an input coarse voxel shape, DECOR-GAN~\cite{chen2021decor} can only adopt the geometric style of one of the reference shapes, while our method enables it to apply different styles to different parts.
}
\label{fig:decorgan}
\end{figure}

DECOR-GAN~\cite{chen2021decor} is a method for conditional voxel upsampling. When trained with $M$ detailed shapes representing $M$ styles, the user can upsample an input coarse voxel grid into a high-resolution, detailed voxel model, but with only one of the $M$ styles learned from the training shapes. The lack of part-level control makes it impossible to combine styles from different shapes, as shown in Figure~\ref{fig:decorgan} (a).

With our predicted segmentation, we provide DECOR-GAN the ability to control per-part geometric styles. In the original DECOR-GAN, each input voxel is associated with one of the $M$ styles. But during training, all voxels from the input shape have to be associated with the same style because there is no part-level segmentation available. We simply modify the training procedure to assign voxels from different parts with different styles. We test our approach on two categories: one is a combined chair+table category, and the other is the plant category. We use the data provided by DECOR-GAN, which are originally from ShapeNet~\cite{chang2015shapenet}. Some qualitative results are shown in Figure~\ref{fig:decorgan}.

\section{Conclusions}
\label{sec:conclusions}

\begin{figure}[t!]
\begin{center}
\includegraphics[width=1.0\linewidth]{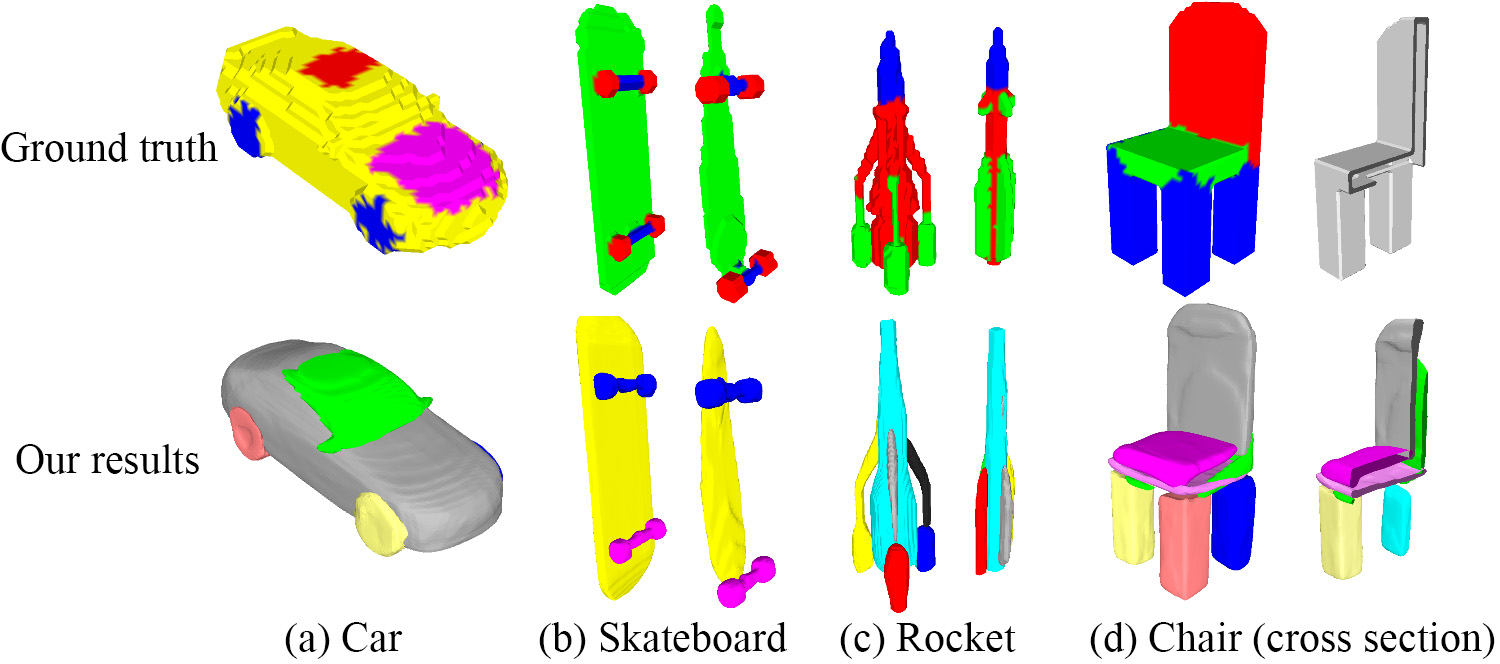}
\end{center}
\caption{
\textbf{Unsuccessful segmentation results}.
}
\label{fig:failure}
\end{figure}

We have introduced the Deforming Auto-Encoder, or DAE-Net, which extends the idea in Transforming Auto-encoders for unsupervised part learning. With our part-wise deformation networks improving the shape reconstruction quality and our training scheme effectively overcoming the local minima issue, our work is the first to show that the TAE framework can yield high-quality, consistent, and fine-grained 3D shape co-segmentation.

Our method segments a shape by reconstructing and deforming individual parts of the shape. As we adopt a volumetric shape representation, our method is unable to segment surface parts, e.g., hood and roof in the car category of ShapeNet; see Figure~\ref{fig:failure} (a).
Likewise, our method may not segment certain semantic parts, as it does not have actual semantic understanding, e.g., the connecting part of the earphone category (blue in ground truth) in Figure~\ref{fig:visual_comparison}, and the wheels (red) and head (blue) of the skateboard and rocket categories in Figures~\ref{fig:failure} (b) and (c), respectively.

Our method also relies heavily on the quality of the training data. Although our training data has been pre-processed to make the voxels as solid as possible, there are still shapes that are hollow inside, e.g., see Figure~\ref{fig:failure} (d).
These hollow shapes will not be correctly segmented, and they may even negatively affect the segmentation quality of other shapes.
On a related matter, parts will not be correctly segmented if they are not correctly reconstructed by our method, either due to underfitting or the parts being too small, as shown in the second rocket example in Figure~\ref{fig:failure} (c).

In the future, we could seek better granularity control by introducing hierarchical structures in the shape representation, as in RIM-Net~\cite{niu2022rim}.
Combining our method with open-vocabulary semantic segmentation methods is also an interesting future direction, which may help produce consistent and meaningful cross-category co-segmentation.

\begin{acks}
We thank all the anonymous reviewers for their insightful comments and constructive feedback. This work was supported in part by an NSERC Discovery Grant (No.~611370) and Adobe gift funds.
\end{acks}

\bibliographystyle{ACM-Reference-Format}
\bibliography{bibliography}


\clearpage

\twocolumn[
\centering
\textbf{\Large\ApproachName{}: Deforming Auto-encoder for unsupervised shape co-segmentation} \\
\vspace{0.5em}\Large{(Supplementary Material)} \\
\vspace{1.0em}
] 
\setcounter{page}{1}

\appendix

\section{Segmentation to skeleton}

\begin{figure}[b!]
\begin{center}
\includegraphics[width=1.0\linewidth]{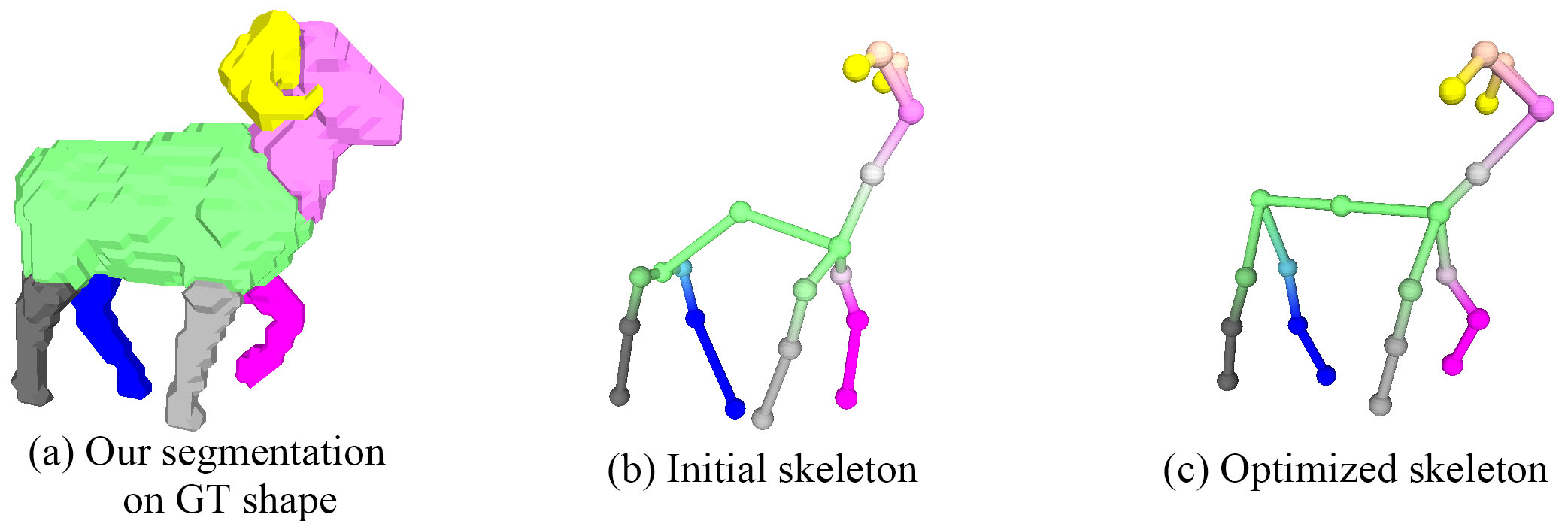}
\end{center}
\caption{
\textbf{Building a skeleton from a segmented shape}.
}
\label{fig:seg2skeleton}
\end{figure}

To build a skeleton, we first use our segmentation method to segment a $64^3$ voxel representation of the shape. We then identify all the parts in this shape. Note that for the same semantic part, we treat each connected component as an individual part, e.g., horns in Figure~\ref{fig:seg2skeleton}. We then create a node at the center of each part, denoted as ``part nodes''. We also create a node whenever two parts meet, and consider these nodes ``joint nodes''. The joint nodes are connected to the corresponding part nodes. We also create additional part nodes according to some heuristics: we create one node to join the front legs, one to join the back legs, and one node at the far end of each limb, as shown in Figure~\ref{fig:seg2skeleton} (b). After creating the initial skeleton, we treat the positions of all part nodes as optimizable parameters, and minimize the symmetric chamfer distance between the points sampled on the skeleton and the points sampled inside the ground truth voxel shape. After optimization with gradient decent, we obtain the final skeleton, as shown in Figure~\ref{fig:seg2skeleton} (c).

\section{More qualitative results}

\begin{figure*}[t!]
\begin{center}
\includegraphics[width=1.0\linewidth]{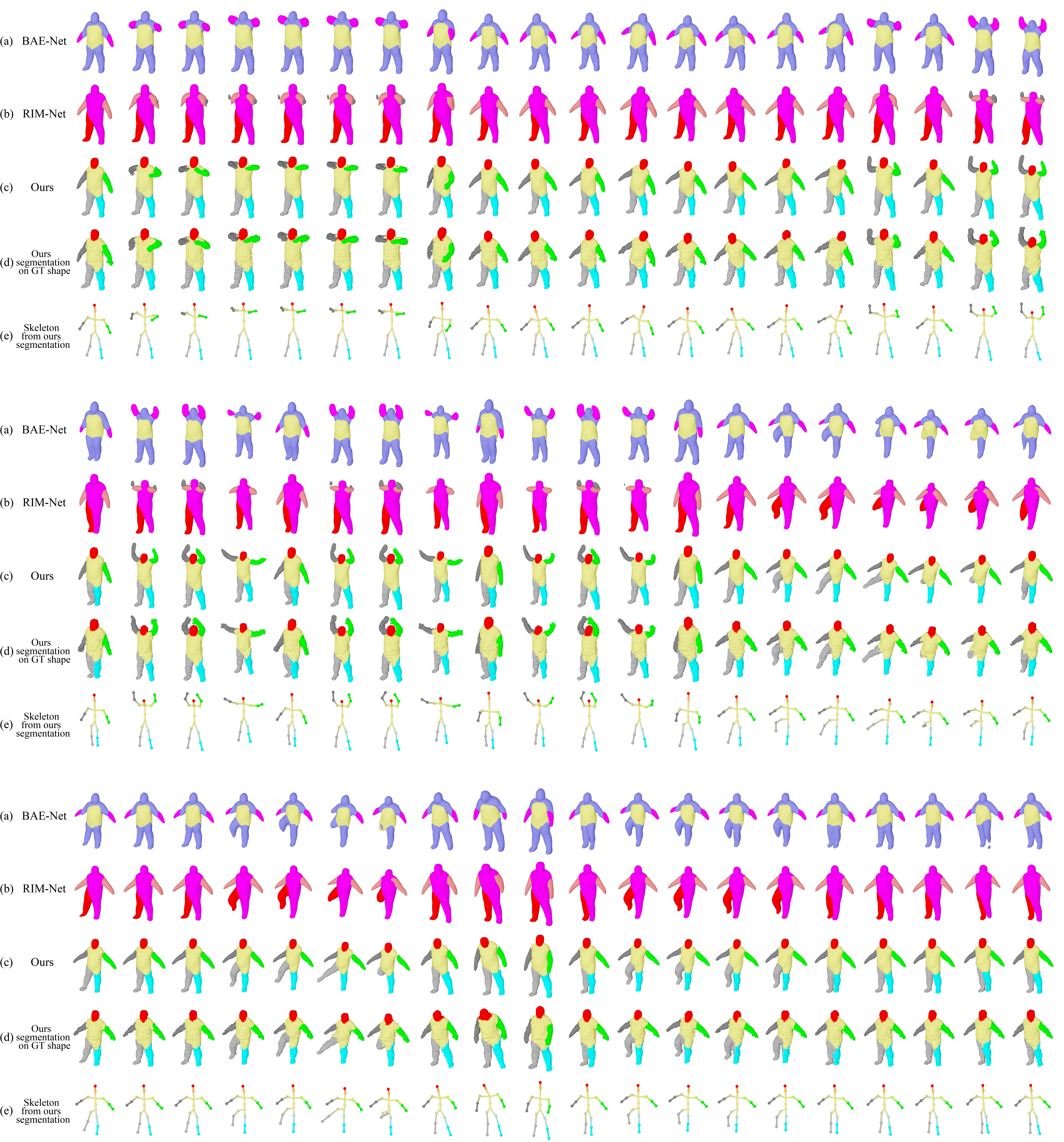}
\end{center}
\caption{
\textbf{Qualitative results on shape segmentation} on \textbf{DFAUST}, compared with other methods.
Within the same category, same color indicates the parts are from the same branch of the network, thus are considered to be corresponded. We also show the skeletons built upon our segmentation.
}
\label{fig:supp_dfaust_1}
\end{figure*}

\begin{figure*}[t!]
\begin{center}
\includegraphics[width=1.0\linewidth]{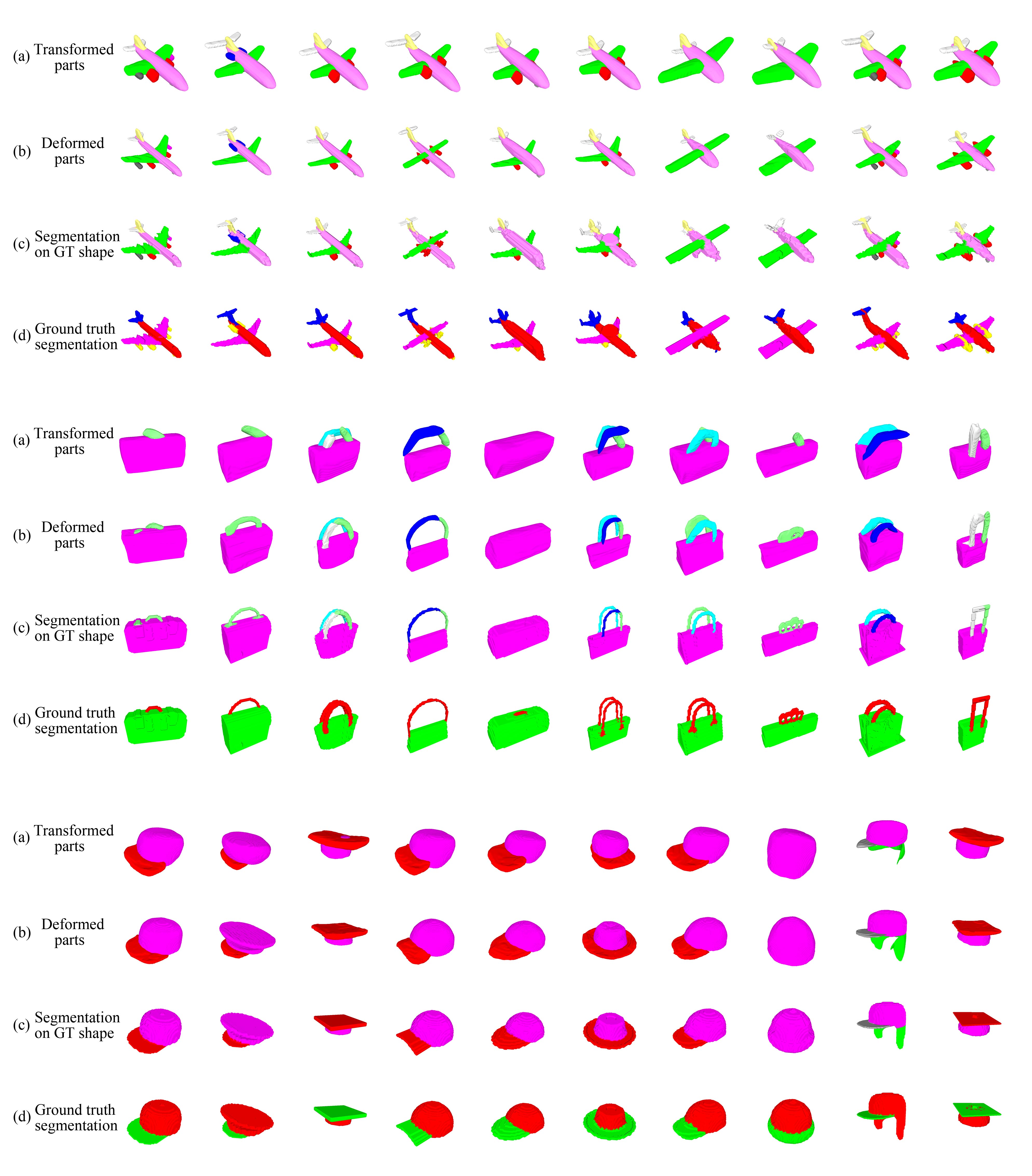}
\end{center}
\caption{
\textbf{Qualitative results on shape segmentation} on the \textbf{airplane}, \textbf{bag}, and \textbf{cap} categories of the ShapeNet Part dataset.
Within the same category, same color indicates the parts are from the same branch of the network, thus are considered to be corresponded.
}
\label{fig:supp_shapenet_1}
\end{figure*}

\begin{figure*}[t!]
\begin{center}
\includegraphics[width=1.0\linewidth]{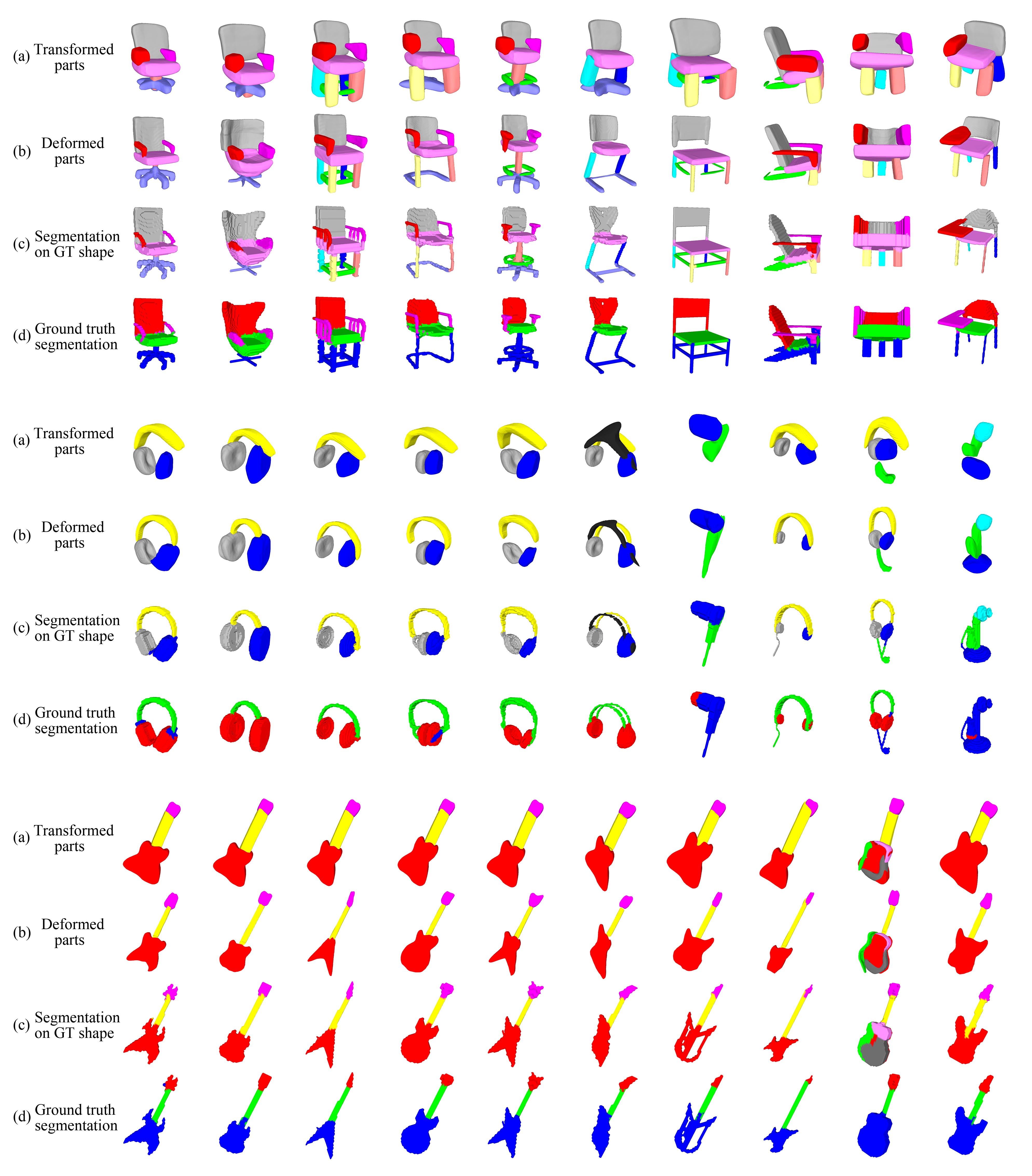}
\end{center}
\caption{
\textbf{Qualitative results on shape segmentation} on the \textbf{chair}, \textbf{earphone}, and \textbf{guitar} categories of the ShapeNet Part dataset.
Within the same category, same color indicates the parts are from the same branch of the network, thus are considered to be corresponded.
}
\label{fig:supp_shapenet_2}
\end{figure*}

\begin{figure*}[t!]
\begin{center}
\includegraphics[width=1.0\linewidth]{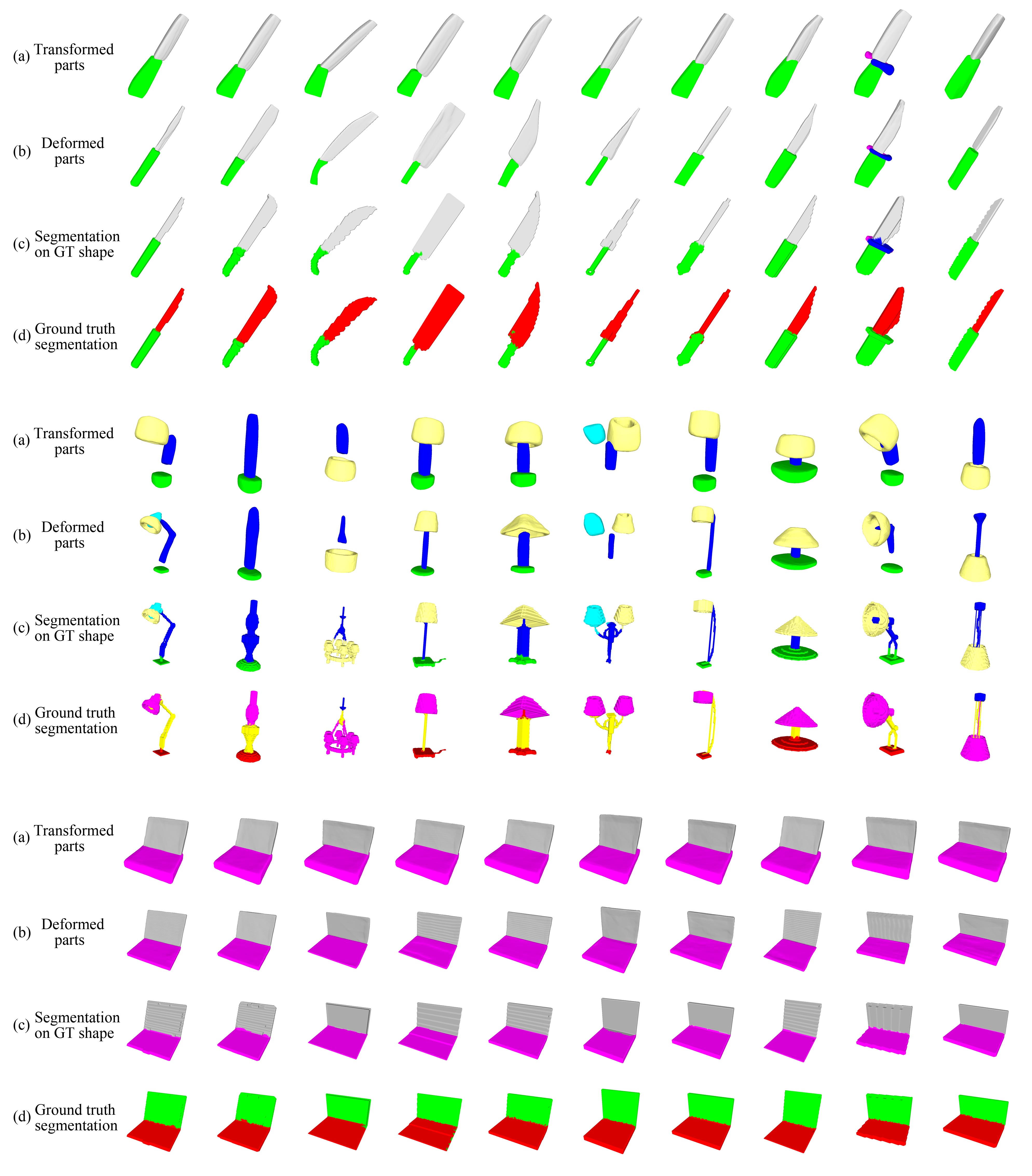}
\end{center}
\caption{
\textbf{Qualitative results on shape segmentation} on the \textbf{knife}, \textbf{lamp}, and \textbf{laptop} categories of the ShapeNet Part dataset.
Within the same category, same color indicates the parts are from the same branch of the network, thus are considered to be corresponded.
}
\label{fig:supp_shapenet_3}
\end{figure*}

\begin{figure*}[t!]
\begin{center}
\includegraphics[width=1.0\linewidth]{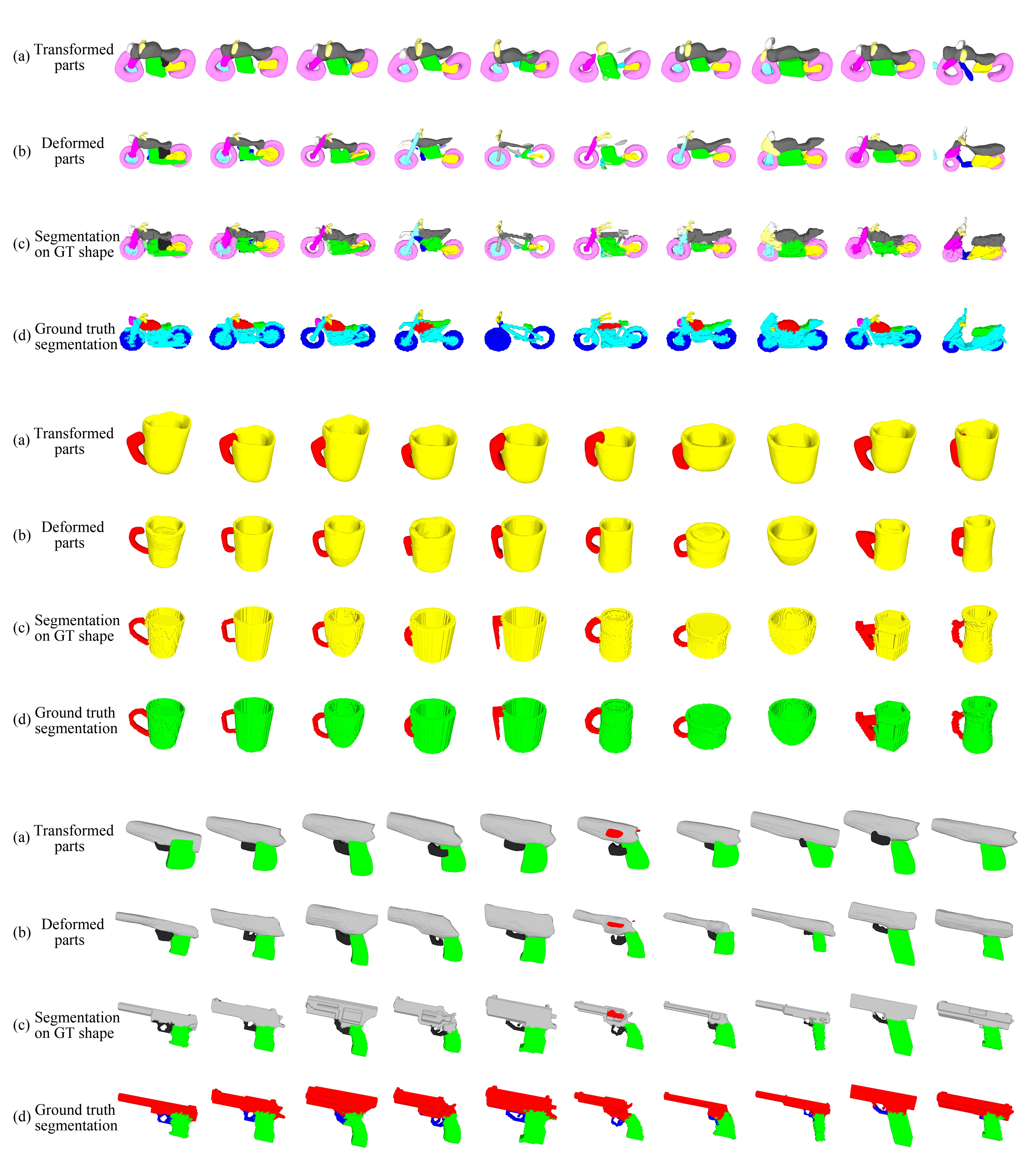}
\end{center}
\caption{
\textbf{Qualitative results on shape segmentation} on the \textbf{motorbike}, \textbf{mug}, and \textbf{pistol} categories of the ShapeNet Part dataset.
Within the same category, same color indicates the parts are from the same branch of the network, thus are considered to be corresponded.
}
\label{fig:supp_shapenet_4}
\end{figure*}

\begin{figure*}[t!]
\begin{center}
\includegraphics[width=1.0\linewidth]{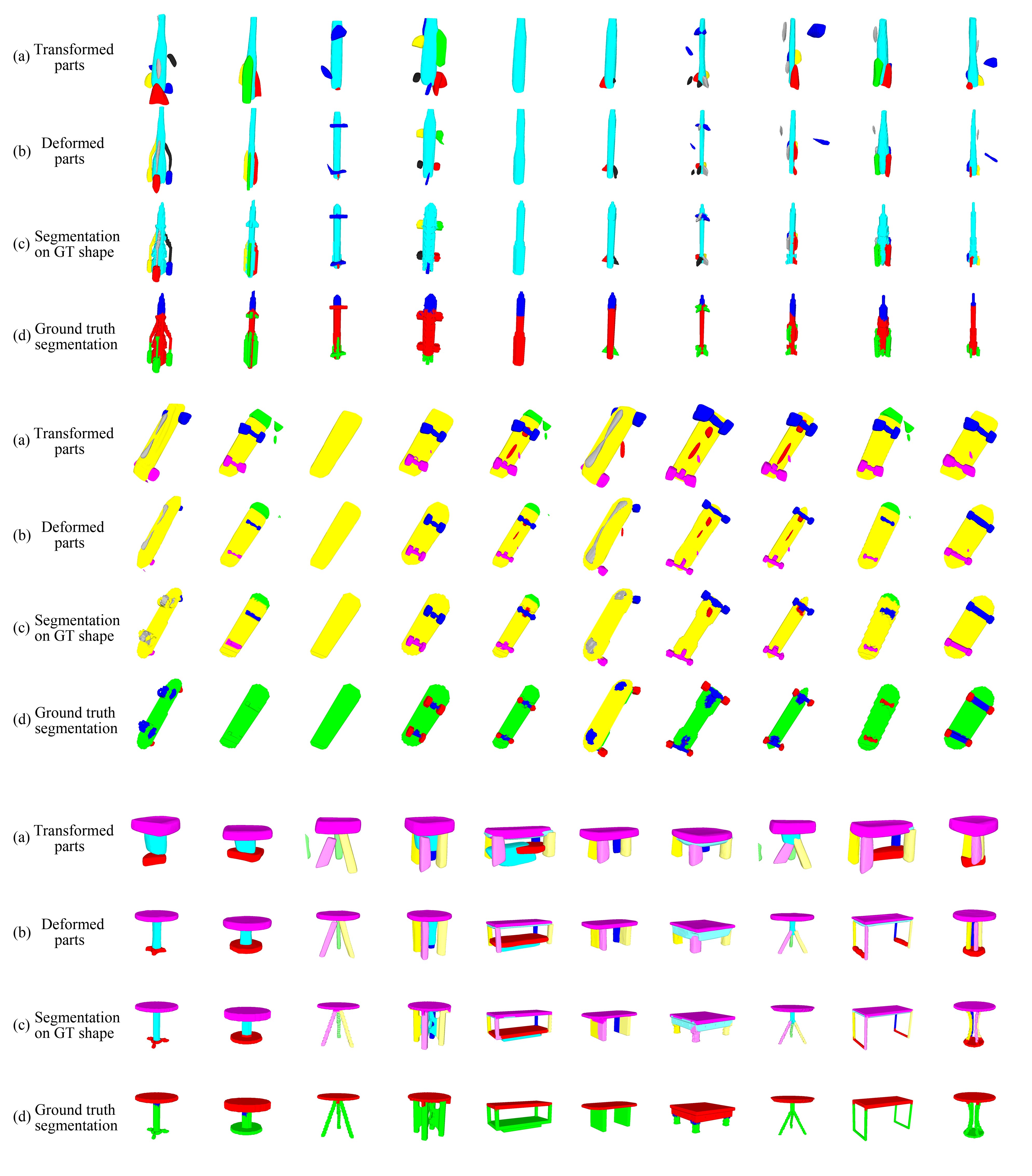}
\end{center}
\caption{
\textbf{Qualitative results on shape segmentation} on the \textbf{rocket}, \textbf{skateboard}, and \textbf{table} categories of the ShapeNet Part dataset.
Within the same category, same color indicates the parts are from the same branch of the network, thus are considered to be corresponded.
}
\label{fig:supp_shapenet_5}
\end{figure*}

We show qualitative results on DFAUST in Figure~\ref{fig:supp_dfaust_1} and ShapeNet Part dataset in Figure~\ref{fig:supp_shapenet_1}~\ref{fig:supp_shapenet_2}~\ref{fig:supp_shapenet_3}~\ref{fig:supp_shapenet_4}~\ref{fig:supp_shapenet_5}.

\clearpage

\clearpage

\end{document}